\documentclass[final]{cvpr}

\usepackage{times}
\usepackage{epsfig}
\usepackage{graphicx}
\usepackage{amsmath}
\usepackage{amssymb}
\usepackage{listings}
\usepackage{booktabs}
\usepackage{float}
\usepackage{bbding}
\usepackage{pifont}
\usepackage{multirow}
\usepackage{textcomp}
\usepackage{caption}
\usepackage{xcolor}
\usepackage[normalem]{ulem}
\usepackage{cuted}

\font\myfont=cmr12 at 15pt
\DeclareMathOperator*{\argmin}{arg\,min}

\usepackage[pagebackref=true,breaklinks=true,colorlinks,bookmarks=false]{hyperref}

\usepackage[pagebackref=true,breaklinks=true,colorlinks,bookmarks=false]{hyperref}
\usepackage{subfiles} 

\def\cvprPaperID{3686} 
\def\confYear{CVPR 2021}

\begin{document}

\title{TableFormer: Table Structure Understanding with Transformers.}

\author{Ahmed Nassar, Nikolaos Livathinos, Maksym Lysak, Peter Staar\\
IBM Research\\
{\tt\small \{ahn,nli,mly,taa\}@zurich.ibm.com }
}

\maketitle

\begin{abstract}

Tables organize valuable content in a concise and compact representation. This content is extremely valuable for systems such as search engines, Knowledge Graph's, etc, since they enhance their predictive capabilities. Unfortunately, tables come in a large variety of shapes and sizes. Furthermore, they can have complex column/row-header configurations, multiline rows, different variety of separation lines, missing entries, etc. As such, the correct identification of the table-structure from an image is a non-trivial task. In this paper, we present a new table-structure identification model. The latter improves the latest end-to-end deep learning model (i.e. encoder-dual-decoder from PubTabNet) in two significant ways. First, we introduce a new object detection decoder for table-cells. In this way, we can obtain the content of the table-cells from programmatic PDF's directly from the PDF source and avoid the training of the custom OCR decoders. This architectural change leads to more accurate table-content extraction and allows us to tackle non-english tables. Second, we replace the LSTM decoders with transformer based decoders. This upgrade improves significantly the previous state-of-the-art tree-editing-distance-score (TEDS) from 91\% to 98.5\% on simple tables and from 88.7\% to 95\% on complex tables.

\end{abstract}

\section{\label{sec:Intro}Introduction}

\begin{figure}[t]
\centering
\includegraphics[scale=0.30]{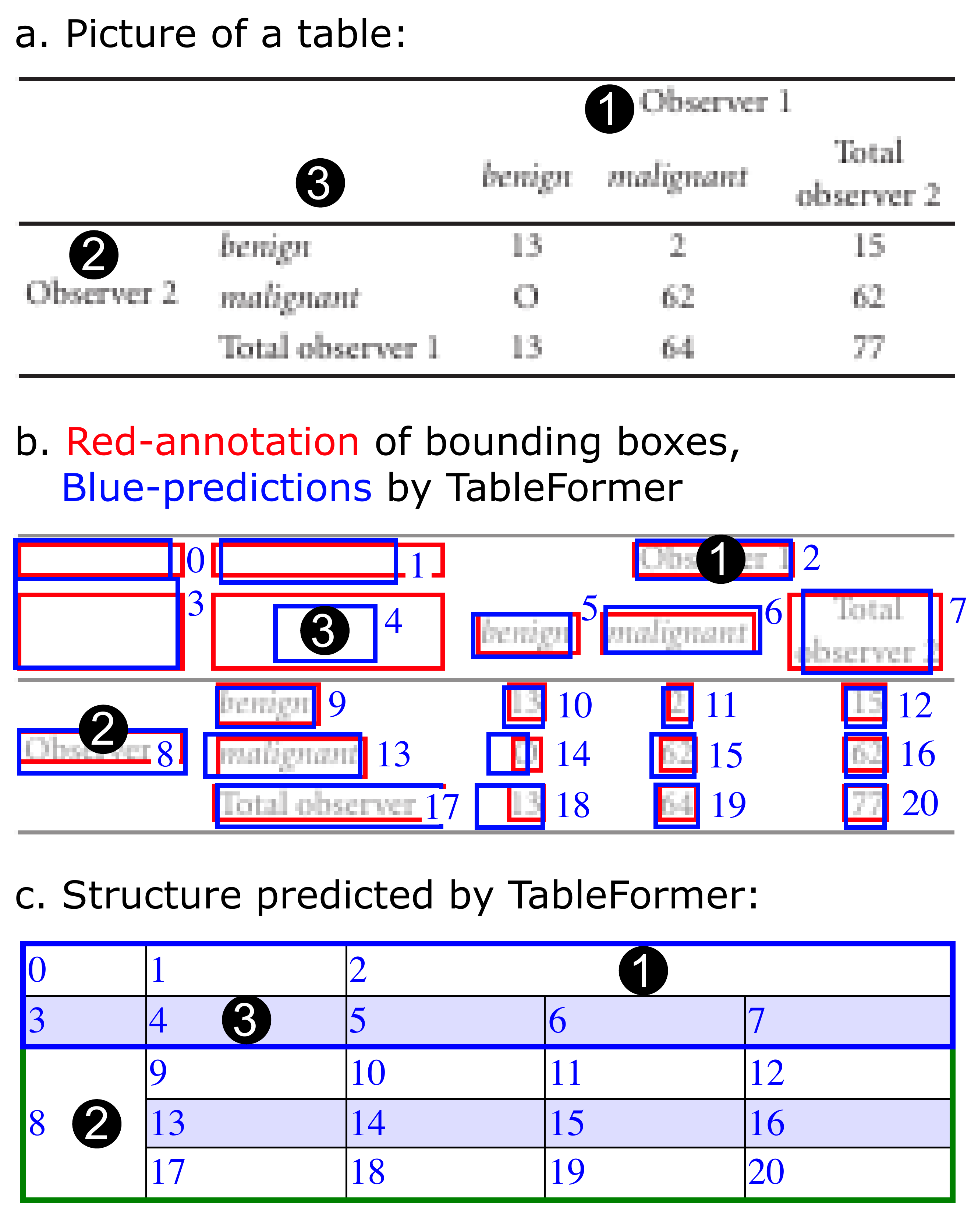}
\caption{\label{fig:Fig1} Picture of a table with subtle, complex features such as (1) multi-column headers, (2) cell with multi-row text and (3) cells with no content. Image from PubTabNet evaluation set, filename: `PMC2944238\_004\_02'.}
\end{figure}

The occurrence of tables in documents is ubiquitous.
They often summarise quantitative or factual data, which is cumbersome to describe in verbose text but nevertheless extremely valuable. Unfortunately, this compact representation is often not easy to parse by machines. There are many implicit conventions used to obtain a compact table representation. For example, tables often have complex column- and row-headers in order to reduce duplicated cell content. Lines of different shapes and sizes are leveraged to separate content or indicate a tree structure. Additionally, tables can also have empty/missing table-entries or multi-row textual table-entries. Fig.~\ref{fig:Fig1} shows a table which presents all these issues.

Recently, significant progress has been made with vision based approaches to extract tables in documents. For the sake of completeness, the issue of table extraction from documents is typically decomposed into two separate challenges, i.e. (1) finding the location of the table(s) on a document-page and (2) finding the structure of a given table in the document.

The first problem is called table-location and has been previously addressed~\cite{KDD2018, PubLayNet, SKI_seq2seq, paliwal2019tablenet, prasad2020cascadetabnet, 8270123, jimaging7100214} with state-of-the-art object-detection networks (e.g. YOLO and later on Mask-RCNN~\cite{MaskRCNN}). For all practical purposes, it can be considered as a solved problem, given enough ground-truth data to train on.

The second problem is called table-structure decomposition. The latter is a long standing problem in the community of document understanding~\cite{ICDAR2013,ICDAR2019,ICDAR2021}. Contrary to the table-location problem, there are no commonly used approaches that can easily be re-purposed to solve this problem. Lately, a set of new model-architectures has been proposed by the community to address table-structure decomposition~\cite{PubTabNet, GTE, GFTE, long2021parsing}. All these models have some weaknesses (see Sec. \ref{sec:SOTA}). The common denominator here is the reliance on textual features and/or the inability to provide the bounding box of each table-cell in the original image.

In this paper, we want to address these weaknesses and present a robust table-structure decomposition algorithm. The design criteria for our model are the following. First, we want our algorithm to be language agnostic. In this way, we can obtain the structure of any table, irregardless of the language. Second, we want our algorithm to leverage as much data as possible from the original PDF document. For programmatic PDF documents, the text-cells can often be extracted much faster and with higher accuracy compared to OCR methods. Last but not least, we want to have a direct link between the table-cell and its bounding box in the image.

To meet the design criteria listed above, we developed a new model called \textbf{TableFormer} and a synthetically generated table structure dataset called \textbf{SynthTabNet}\footnote{https://github.com/IBM/SynthTabNet}. In particular, our contributions in this work can be summarised as follows:
\begin{itemize}
  \item We propose \textbf{TableFormer}, a transformer based model that predicts tables structure and bounding boxes for the table content simultaneously in an end-to-end approach.
  \item Across all benchmark datasets \textbf{TableFormer} significantly outperforms existing state-of-the-art metrics, while being much more efficient in training and inference to existing works.
  \item We present \textbf{SynthTabNet} a synthetically generated dataset, with various appearance styles and complexity.
  \item An augmented dataset based on PubTabNet~\cite{PubTabNet}, FinTabNet~\cite{GTE}, and TableBank~\cite{tablebank} with generated ground-truth for reproducibility.
\end{itemize}


The paper is structured as follows. In Sec.~\ref{sec:SOTA}, we give a brief overview of the current state-of-the-art. In Sec.~\ref{sec:Data}, we describe the datasets on which we train. In Sec.~\ref{sec:models}, we introduce the TableFormer model-architecture and describe its results \& performance in Sec.~\ref{sec:results}. As a conclusion, we describe how this new model-architecture can be re-purposed for other tasks in the computer-vision community.

\section{\label{sec:SOTA}Previous work and State of the Art}


Identifying the structure of a table has been an outstanding problem in the document-parsing community, that motivates many organised public challenges~\cite{ICDAR2013, ICDAR2019, ICDAR2021}. The difficulty of the problem can be attributed to a number of factors. First, there is a large variety in the shapes and sizes of tables. Such large variety requires a flexible method. This is especially true for complex column- and row headers, which can be extremely intricate and demanding. A second factor of complexity is the lack of data with regard to table-structure. Until the publication of PubTabNet~\cite{PubTabNet}, there were no large datasets (i.e. $>100$K tables) that provided structure information. This happens primarily due to the fact that tables are notoriously time-consuming to annotate by hand. However, this has definitely changed in recent years with the deliverance of PubTabNet~\cite{PubTabNet}, FinTabNet~\cite{GTE}, TableBank~\cite{tablebank} etc.


Before the rising popularity of deep neural networks, the community relied heavily on heuristic and/or statistical methods to do table structure identification \cite{Couasnon2014, green95recognition, hu1999medium, gatos2005automatic, kasar2013learning, shafait2010table}.
Although such methods work well on constrained tables~\cite{constrained_table_structure}, a more data-driven approach can be applied due to the advent of convolutional neural networks (CNNs) and the availability of large datasets.
To the best-of-our knowledge, there are currently two different types of network architecture that are being pursued for state-of-the-art table-structure identification.


\textbf{Image-to-Text networks}: In this type of network, one predicts a sequence of tokens starting from an encoded image. Such sequences of tokens can be HTML table tags~\cite{PubTabNet, tablebank} or LaTeX symbols\cite{He2021PingAnVCGroupsSF}. The choice of symbols is ultimately not very important, since one can be transformed into the other. There are however subtle variations in the Image-to-Text networks. The easiest network architectures are ``image-encoder $\rightarrow$ text-decoder'' (IETD), similar to network architectures that try to provide captions to images~\cite{ShowAndTell}. In these IETD networks, one expects as output the LaTeX/HTML string of the entire table, i.e. the symbols necessary for creating the table with the content of the table. Another approach is the ``image-encoder $\rightarrow$ dual decoder'' (IEDD) networks. In these type of networks, one has two consecutive decoders with different purposes. The first decoder is the \textit{tag-decoder}, i.e. it only produces the HTML/LaTeX tags which construct an empty table. The second \textit{content-decoder} uses the encoding of the image in combination with the output encoding of each cell-tag (from the \textit{tag-decoder}) to generate the textual content of each table cell. The network architecture of IEDD is certainly more elaborate, but it has the advantage that one can pre-train the tag-decoder which is constrained to the table-tags. 


In practice, both network architectures (IETD and IEDD) require an implicit, custom trained object-character-recognition (OCR) to obtain the content of the table-cells. In the case of IETD, this OCR engine is implicit in the decoder similar to \cite{qasim2019rethinking}. For the IEDD, the OCR is solely embedded in the content-decoder. This reliance on a custom, implicit OCR decoder is of course problematic. OCR is a well known and extremely tough problem, that often needs custom training for each individual language.
However, the limited availability for non-english content in the current datasets, makes it impractical to apply the IETD and IEDD methods on tables with other languages. Additionally, OCR can be completely omitted if the tables originate from programmatic PDF documents with known positions of each cell. The latter was the inspiration for the work of this paper.

\textbf{Graph Neural networks}: Graph Neural networks (GNN's) take a radically different approach to table-structure extraction. Note that one table cell can constitute out of multiple text-cells. To obtain the table-structure, one creates an initial graph, where each of the text-cells becomes a node in the graph similar to \cite{xue2019res2tim, xue2021tgrnet, chi2019complicated}. Each node is then associated with en embedding vector coming from the encoded image, its coordinates and the encoded text. Furthermore, nodes that represent adjacent text-cells are linked. Graph Convolutional Networks (GCN's) based methods take the image as an input, but also the position of the text-cells and their content~\cite{GFTE}. The purpose of a GCN is  to transform the input graph into a new graph, which replaces the old links with new ones. The new links then represent the table-structure. With this approach, one can avoid the need to build custom OCR decoders. However, the quality of the reconstructed structure is not comparable to the current state-of-the-art~\cite{GFTE}.

\textbf{Hybrid Deep Learning-Rule-Based approach}: A popular current model for table-structure identification is the use of a hybrid Deep Learning-Rule-Based approach similar to \cite{schreiber2017deepdesrt, siddiqui2019deeptabstr}. In this approach, one first detects the position of the table-cells with object detection (e.g. YoloVx or Mask-RCNN), then classifies the table into different types (from its images) and finally uses different rule-sets to obtain its table-structure. Currently, this approach achieves state-of-the-art results, but is not an end-to-end deep-learning method. As such, new rules need to be written if different types of tables are encountered. 

\begin{figure}[t!]\centering
	\includegraphics[scale=0.25]{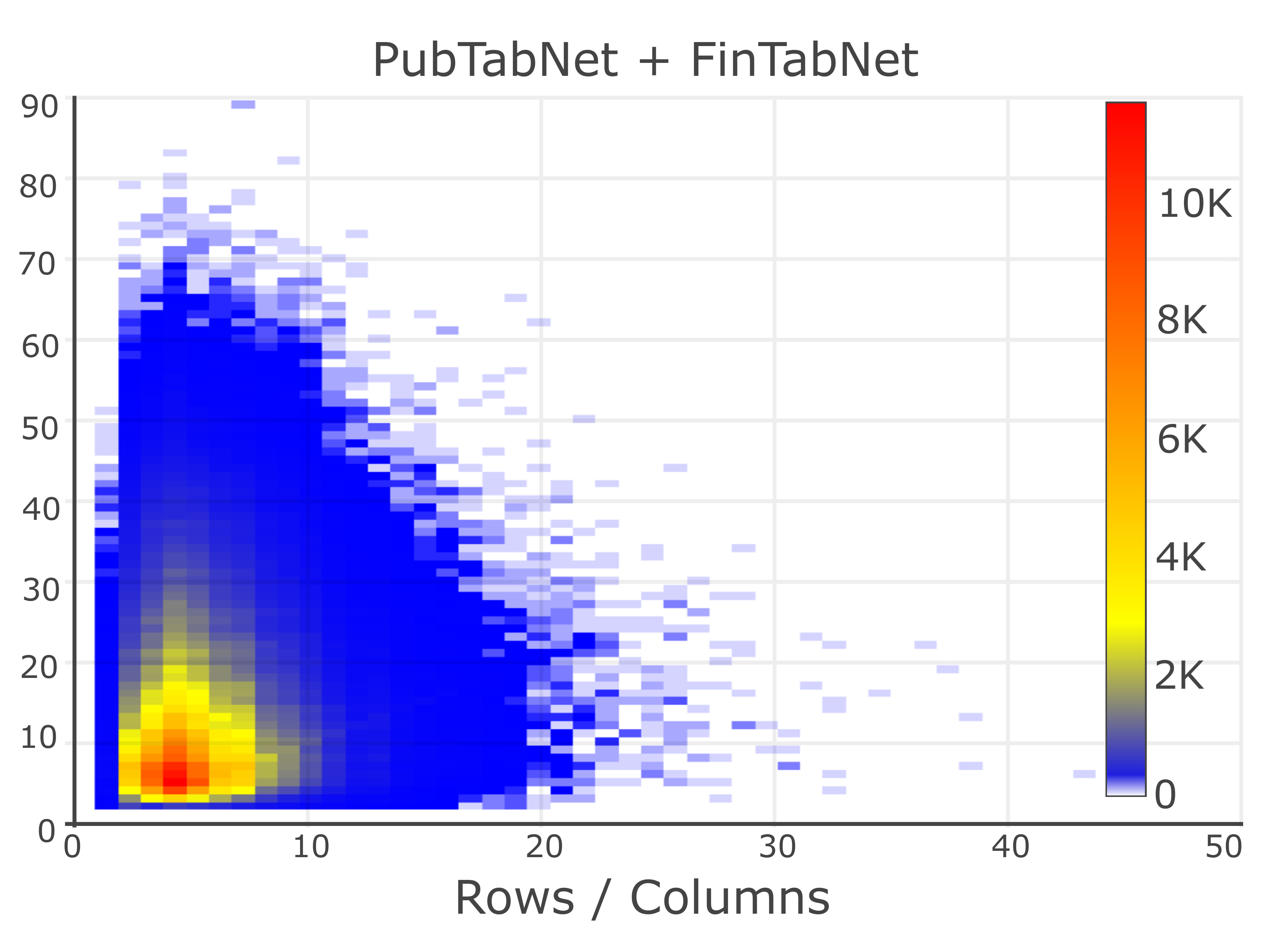}
	\caption{\label{fig:Fig2} Distribution of the tables across different table dimensions in PubTabNet + FinTabNet datasets}
\end{figure}

\section{\label{sec:Data}Datasets}


We rely on large-scale datasets such as PubTabNet~\cite{PubTabNet}, FinTabNet~\cite{GTE}, and TableBank~\cite{tablebank} datasets to train and evaluate our models. These datasets span over various appearance styles and content. We also introduce our own synthetically generated SynthTabNet dataset to fix an imbalance in the previous datasets.

The PubTabNet dataset contains 509k tables delivered as annotated PNG images. The annotations consist of the table structure represented in HTML format, the tokenized text and its bounding boxes per table cell. Fig. \ref{fig:Fig1} shows the appearance style of PubTabNet. Depending on its complexity, a table is characterized as ``simple'' when it does not contain row spans or column spans, otherwise it is ``complex''. The dataset is divided into Train and Val splits (roughly 98\% and 2\%). The Train split consists of 54\% simple and 46\% complex tables and the Val split of 51\% and 49\% respectively. The FinTabNet dataset contains 112k tables delivered as single-page PDF documents with mixed table structures and text content. Similarly to the PubTabNet, the annotations of FinTabNet include the table structure in HTML, the tokenized text and the bounding boxes on a table cell basis. The dataset is divided into Train, Test and Val splits (81\%, 9.5\%, 9.5\%), and each one is almost equally divided into simple and complex tables (Train: 48\% simple, 52\% complex, Test: 48\% simple, 52\% complex, Test: 53\% simple, 47\% complex). Finally the TableBank dataset consists of 145k tables provided as JPEG images. The latter has annotations for the table structure, but only few with bounding boxes of the table cells. The entire dataset consists of simple tables and it is divided into 90\% Train, 3\% Test and 7\% Val splits.

Due to the heterogeneity across the dataset formats, it was necessary to combine all available data into one homogenized dataset before we could train our models for practical purposes. Given the size of PubTabNet, we adopted its annotation format and we extracted and converted all tables as PNG images with a resolution of 72 dpi. Additionally, we have filtered out tables with extreme sizes due to small amount of such tables, and kept only those ones ranging between 1*1 and 20*10 (rows/columns).


The availability of the bounding boxes for all table cells is essential to train our models. In order to distinguish between empty and non-empty bounding boxes, we have introduced a binary class in the annotation.
Unfortunately, the original datasets either omit the bounding boxes for whole tables (e.g. TableBank) or they narrow their scope only to non-empty cells. Therefore, it was imperative to introduce a data pre-processing procedure that generates the missing bounding boxes out of the annotation information. This procedure first parses the provided table structure and calculates the dimensions of the most fine-grained grid that covers the table structure. Notice that each table cell may occupy multiple grid squares due to row or column spans. In case of PubTabNet we had to compute missing bounding boxes for 48\% of the simple and 69\% of the complex tables. Regarding FinTabNet, 68\% of the simple and 98\% of the complex tables require the generation of bounding boxes.

As it is illustrated in Fig.~\ref{fig:Fig2}, the table distributions from all datasets are skewed towards simpler structures with fewer number of rows/columns.
Additionally, there is very limited variance in the table styles, which in case of PubTabNet and FinTabNet means one styling format for the majority of the tables.
Similar limitations appear also in the type of table content, which in some cases (e.g. FinTabNet) is restricted to a certain domain. Ultimately, the lack of diversity in the training dataset damages the ability of the models to generalize well on unseen data.

Motivated by those observations we aimed at generating a synthetic table dataset named \textit{SynthTabNet}. This approach offers control over: 1) the size of the dataset, 2) the table structure, 3) the table style and 4) the type of content. The complexity of the table structure is described by the size of the table header and the table body, as well as the percentage of the table cells covered by row spans and column spans. A set of carefully designed styling templates provides the basis to build a wide range of table appearances. Lastly, the table content is generated out of a curated collection of text corpora. By controlling the size and scope of the synthetic datasets we are able to train and evaluate our models in a variety of different conditions.
For example, we can first generate a highly diverse dataset to train our models and then evaluate their performance on other synthetic datasets which are focused on a specific domain.

In this regard, we have prepared four synthetic datasets, each one containing 150k examples. The corpora to generate the table text consists of the most frequent terms appearing in PubTabNet and FinTabNet together with randomly generated text. The first two synthetic datasets have been fine-tuned to mimic the appearance of the original datasets but encompass more complicated table structures. The third one adopts a colorful appearance with high contrast and the last one contains tables with sparse content. Lastly, we have combined all synthetic datasets into one big unified synthetic dataset of 600k examples.

Tab.~\ref{tab:datasets} summarizes the various attributes of the datasets.

\begin{table}[t]\centering
\begin{tabular}{l|cccccc}
          & Tags & Bbox & Size & Format \\ \hline 
PubTabNet & \textcolor{black}{\ding{51}} & \textcolor{black}{\ding{51}} & 509k  & PNG \\
FinTabNet & \textcolor{black}{\ding{51}} & \textcolor{black}{\ding{51}} & 112k & PDF \\
TableBank & \textcolor{black}{\ding{51}} & \textcolor{red}{\ding{55}} & 145k & JPEG \\
Combined-Tabnet(*) & \textcolor{black}{\ding{51}} & \textcolor{black}{\ding{51}} & 400k & PNG \\
Combined(**) & \textcolor{black}{\ding{51}} & \textcolor{black}{\ding{51}} & 500k & PNG \\
SynthTabNet & \textcolor{black}{\ding{51}} & \textcolor{black}{\ding{51}} & 600k & PNG\\

\end{tabular}
\caption{\label{tab:datasets} Both \textit{``Combined-Tabnet"} and \textit{"Combined-Tabnet"} are variations of the following: (*) The Combined-Tabnet dataset is the processed combination of PubTabNet and Fintabnet. (**) The combined dataset is the processed combination of PubTabNet, Fintabnet and TableBank.}
\end{table}

\begin{figure*}[ht!]\centering
\centering
\includegraphics[scale=0.17]{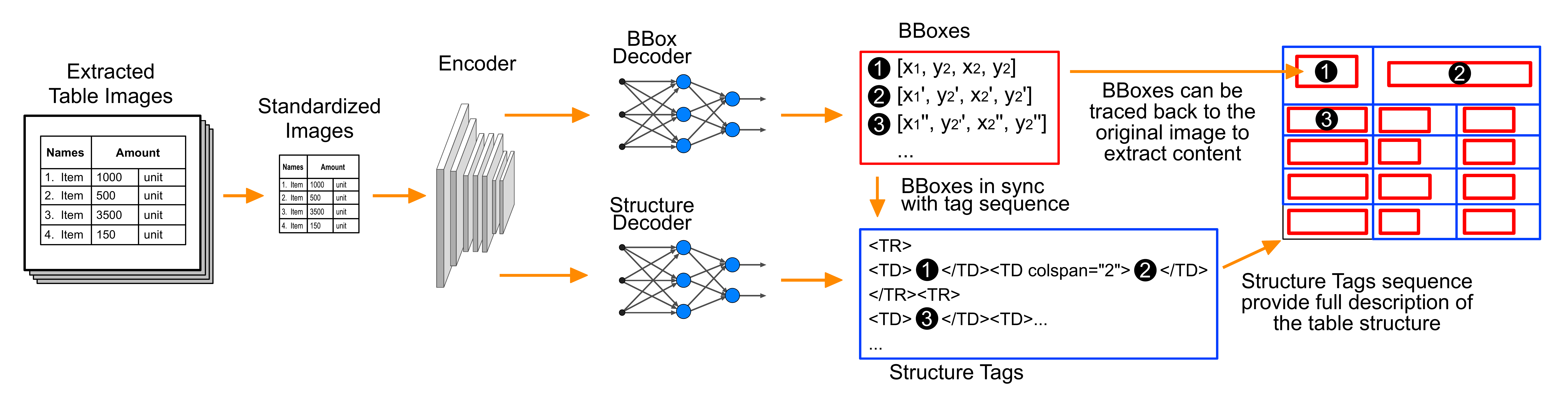}
    \caption{\label{fig:Fig3} \textbf{TableFormer} takes in an image of the PDF and creates bounding box and HTML structure predictions that are synchronized. The bounding boxes grabs the content from the PDF and inserts it in the structure.}
\end{figure*}


%
\section{\label{sec:models} The TableFormer model}

Given the image of a table, TableFormer is able to predict:
1) a sequence of tokens that represent the structure of a table, and
2) a bounding box coupled to a subset of those tokens.
The conversion of an image into a sequence of tokens is a well-known task~\cite{you2016image, 6522402}. While attention is often used as an implicit method to associate each token of the sequence with a position in the original image,
an explicit association between the individual table-cells and the image bounding boxes is also required.

\subsection{\label{sec:eddv2} Model architecture.}
We now describe in detail the proposed method, which is composed of three main components, see Fig. \ref{fig:tablemodel04_diagram}. 
Our \textit{CNN Backbone Network} encodes the input as a feature vector of predefined length.
The input feature vector of the encoded image is passed to the \textit{Structure Decoder} to produce a sequence of HTML tags that represent the structure of the table.
With each prediction of an HTML standard data cell (`\lstinline|<td>|') the hidden state of that cell is passed to the Cell BBox Decoder.
As for spanning cells, such as row or column span, the tag is broken down to `\lstinline|<|', `\lstinline|rowspan=|' or `\lstinline|colspan=|', with the number of spanning cells (attribute), and `\lstinline|>|'. The hidden state attached to `\lstinline|<|' is passed to the Cell BBox Decoder. A shared feed forward network (FFN) receives the hidden states from the Structure Decoder, to provide the final detection predictions of the bounding box coordinates and their classification.

\label{fig:tablemodel04_diagram}
\begin{figure}
\centering
\includegraphics[scale=0.25]{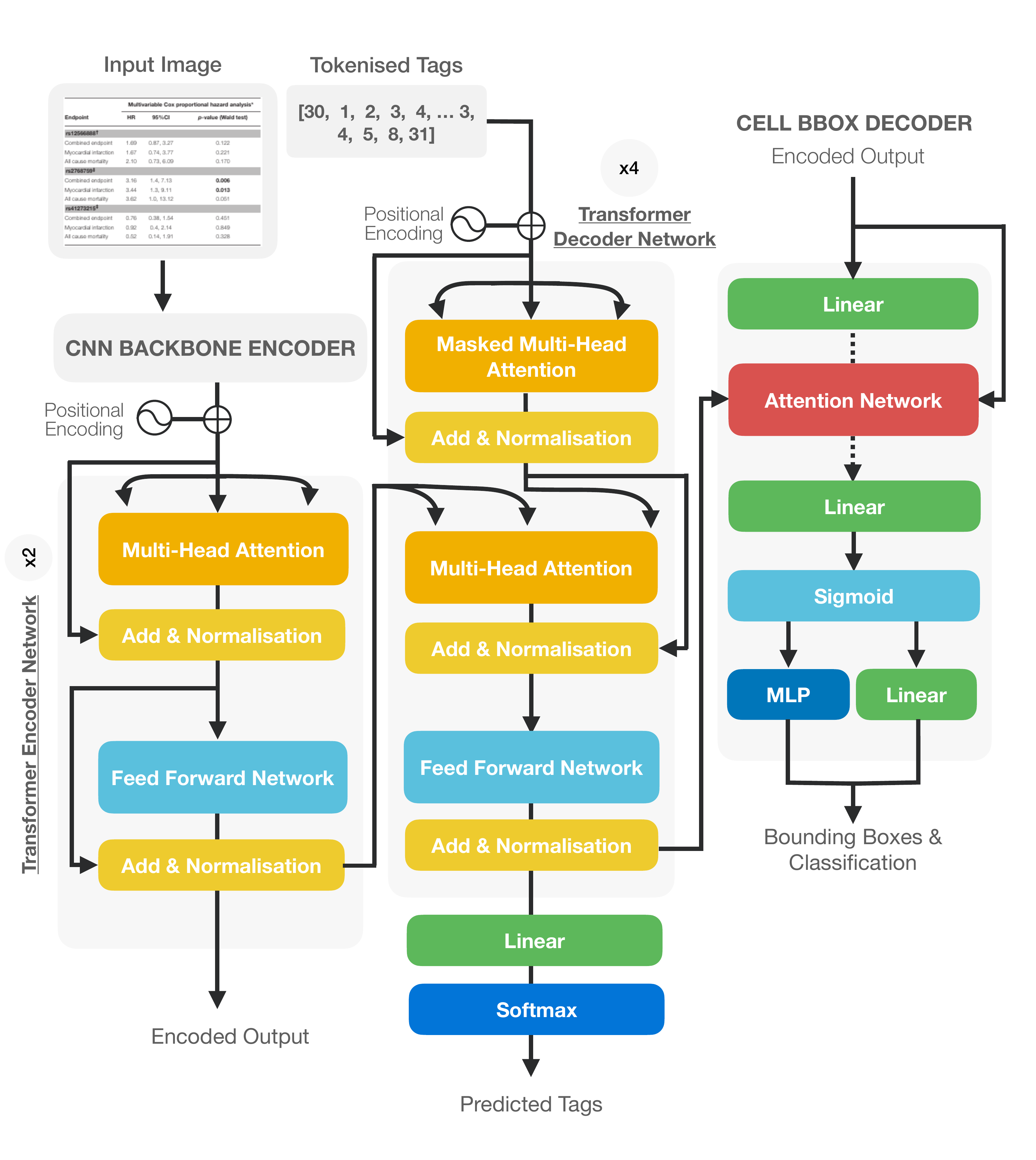}
\caption{\label{fig:tablemodel04_diagram} Given an input image of a table, the \textbf{Encoder} produces fixed-length features that represent the input image.
The features are then passed to both the \textbf{Structure Decoder} and \textbf{Cell BBox Decoder}.
During training, the \textbf{Structure Decoder} receives `tokenized tags' of the HTML code that represent the table structure.
Afterwards, a transformer encoder and decoder architecture is employed to produce features that are received by a linear layer, and the \textbf{Cell BBox Decoder. The linear layer is applied to the features to predict the tags. Simultaneously, the} \textbf{Cell BBox Decoder} selects features referring to the data cells (`\lstinline|<td>|', `\lstinline|<|') and passes them through an attention network, an MLP, and a linear layer to predict the bounding boxes.}
\end{figure}

\textbf{CNN Backbone Network.} A ResNet-18 CNN is the backbone that receives the table image and encodes it as a vector of predefined length.
The network has been modified by removing the linear and pooling layer, as we are not performing classification, and adding an adaptive pooling layer of size 28*28. ResNet by default downsamples the image resolution by 32 and then the encoded image is provided to both the \textit{Structure Decoder}, and \textit{Cell BBox Decoder}.

\textbf{Structure Decoder.} The transformer architecture of this component is based on the work proposed in~\cite{NIPS2017_7181}. After extensive experimentation, the \textit{Structure Decoder} is modeled as a transformer encoder with two encoder layers and a transformer decoder made from a stack of 4 decoder layers that comprise mainly of multi-head attention and feed forward layers. This configuration uses fewer layers and heads in comparison to networks applied to other problems (e.g. ``Scene Understanding'', ``Image Captioning''), something which we relate to the simplicity of table images.

The transformer encoder receives an encoded image from the \textit{CNN Backbone Network} and refines it through a multi-head dot-product attention layer, followed by a Feed Forward Network.
During training, the transformer decoder receives as input the output feature produced by the transformer encoder, and the tokenized input of the HTML ground-truth tags. Using a stack of multi-head attention layers, different aspects of the tag sequence could be inferred. This is achieved by each attention head on a layer operating in a different subspace, and then combining altogether their attention score.

\textbf{Cell BBox Decoder.} Our architecture allows to simultaneously predict HTML tags and bounding boxes for each table cell without the need of a separate object detector end to end.
This approach is inspired by DETR \cite{10.1007/978-3-030-58452-8_13} which employs a Transformer Encoder, and Decoder that looks for a specific number of object queries (potential object detections).
As our model utilizes a transformer architecture, the hidden state of the \lstinline|<td>|' and `\lstinline|<|' HTML structure tags become the object query.

The encoding generated by the \textit{CNN Backbone Network} along with the features acquired for every data cell from the Transformer Decoder are then passed to the attention network. The attention network takes both inputs and learns to provide an attention weighted encoding. This weighted attention encoding is then multiplied to the encoded image to produce a feature for each table cell.
Notice that this is different than the typical object detection problem where imbalances between the number of detections and the amount of objects may exist. In our case, we know up front that the produced detections always match with the table cells in number and correspondence.

The output features for each table cell are then fed into the feed-forward network (FFN).
The FFN consists of a Multi-Layer Perceptron (3 layers with ReLU activation function) that predicts the normalized coordinates for the bounding box of each table cell.
Finally, the predicted bounding boxes are classified based on whether they are empty or not using a linear layer.

\textbf{Loss Functions.} We formulate a multi-task loss Eq. \ref{eq:loss} to train our network.
The Cross-Entropy loss (denoted as $l_{s}$) is used to train the \textit{Structure Decoder} which predicts the structure tokens.
As for the \textit{Cell BBox Decoder} it is trained with a combination of losses denoted as $l_{box}$. $l_{box}$ consists of the generally used $l_{1}$ loss for object detection and the IoU loss ($l_{iou}$) to be scale invariant as explained in \cite{rezatofighi2019generalized}. In comparison to DETR, we do not use the Hungarian algorithm \cite{kuhn1955hungarian} to match the predicted bounding boxes with the ground-truth boxes, as we have already achieved a one-to-one match through two steps:
1) Our token input sequence is naturally ordered, therefore the hidden states of the table data cells are also in order when they are provided as input to the \textit{Cell BBox Decoder}, and
2) Our bounding boxes generation mechanism (see Sec. \ref{sec:Data}) ensures a one-to-one mapping between the cell content and its bounding box for all post-processed datasets.

	The loss used to train the TableFormer can be defined as following:

	\label{eq:loss}
	\begin{equation}
	\begin{split}
	l_{box} = \lambda_{iou}l_{iou}+\lambda_{l1} \\
  	l = \lambda l_{s}+(1-\lambda) l_{box}
  	\end{split}
	\end{equation}
	where $\lambda \in$ [0, 1], and $\lambda_{iou}, \lambda_{l1} \in \Bbb{R}$ are hyper-parameters.

\section{\label{sec:results}Experimental Results}
\subsection{\label{sec:implementation_details}Implementation Details}
TableFormer uses ResNet-18 as the \textit{CNN Backbone Network}. The input images are resized to 448*448 pixels and the feature map has a dimension of 28*28.
Additionally, we enforce the following input constraints:
	\begin{equation}
	\begin{split}
	\textnormal{Image width and height $ \le $ 1024 pixels} \\
	\textnormal{Structural tags length $ \le $ 512 tokens.}
	\end{split}
	\end{equation}
Although input constraints are used also by other methods, such as EDD, ours are less restrictive due to the improved runtime performance and lower memory footprint of TableFormer. This allows to utilize input samples with longer sequences and images with larger dimensions.

The Transformer Encoder consists of two ``Transformer Encoder Layers",  with an input feature size of 512, feed forward network of 1024, and 4 attention heads. As for the Transformer Decoder it is composed of four ``Transformer Decoder Layers" with similar input and output dimensions as the ``Transformer Encoder Layers".
Even though our model uses fewer layers and heads than the default implementation parameters, our extensive experimentation has proved this setup to be more suitable for table images. We attribute this finding to the inherent design of table images, which contain mostly lines and text, unlike the more elaborate content present in other scopes (e.g. the COCO dataset).
Moreover, we have added ResNet blocks to the inputs of the Structure Decoder and Cell BBox Decoder.
This prevents a decoder having a stronger influence over the learned weights which would damage the other prediction task (structure vs bounding boxes), but learn task specific weights instead.
Lastly our dropout layers are set to 0.5.

For training, TableFormer is trained with 3 Adam optimizers, each one for the \textit{CNN Backbone Network}, \textit{Structure Decoder}, and \textit{Cell BBox Decoder}. Taking the PubTabNet as an example for our parameter set up, the initializing learning rate is 0.001 for 12 epochs with a batch size of 24, and $\lambda$ set to 0.5.
Afterwards, we reduce the learning rate to 0.0001, the batch size to 18 and train for 12 more epochs or convergence.

TableFormer is implemented with PyTorch and Torchvision libraries \cite{NEURIPS2019_9015}.
To speed up the inference, the image undergoes a single forward pass through the \textit{CNN Backbone Network} and transformer encoder. This eliminates the overhead of generating the same features for each decoding step.
Similarly, we employ a 'caching' technique to preform faster autoregressive decoding. This is achieved by storing the features of decoded tokens so we can reuse them for each time step. Therefore, we only compute the attention for each new tag.

\subsection{Generalization}
TableFormer is evaluated on three major publicly available datasets of different nature to prove the generalization and effectiveness of our model.
The datasets used for evaluation are the PubTabNet, FinTabNet and TableBank which stem from the scientific, financial and general domains respectively.

We also share our baseline results on the challenging SynthTabNet dataset. Throughout our experiments, the same parameters stated in Sec. \ref{sec:implementation_details} are utilized.

\subsection{Datasets and Metrics}
The Tree-Edit-Distance-Based Similarity (TEDS) metric was introduced in \cite{PubTabNet}. It represents the prediction, and ground-truth as a tree structure of HTML tags. This similarity is calculated as:

\begin{equation}
  \operatorname{TEDS}\left(T_{a}, T_{b}\right)=1-\frac{\operatorname{EditDist}\left(T_{a}, T_{b}\right)}{\max \left(\left|T_{a}\right|,\left|T_{b}\right|\right)}
\end{equation}

where $T_{a}$ and $T_{b}$ represent tables in tree structure HTML format. EditDist denotes the tree-edit distance, and $|T|$ represents the number of nodes in $T$.
\subsection{Quantitative Analysis}
\textbf{Structure.} As shown in Tab. \ref{tab:structure_results}, TableFormer outperforms all SOTA methods across different datasets by a large margin for predicting the table structure from an image.
All the more, our model outperforms pre-trained methods.
During the evaluation we do not apply any table filtering.
We also provide our baseline results on the SynthTabNet  dataset. It has been observed that large tables (e.g. tables that occupy half of the page or more) yield poor predictions. We attribute this issue to the image resizing during the pre-processing step, that produces downsampled images with indistinguishable features.
This problem can be addressed by treating such big tables with a separate model which accepts a large input image size.

\begin{table}[H]
\centering
\begin{tabular}{@{}cccccc@{}}
\toprule
& \multirow{2}{*}{Model} & & \multicolumn{3}{c}{TEDS} \\
& & Dataset & Simple & Complex  & All \\ \midrule
& EDD & PTN & 91.1 & 88.7 & 89.9 \\
& GTE  & PTN & - & - & 93.01 \\ \rule{0pt}{2ex}    

& TableFormer & PTN & 98.5 & 95.0 & \textbf{96.75}  \\ 
\hline \rule{0pt}{2.5ex}
& EDD & FTN &  88.4 &  92.08 &  90.6 \\ 
& GTE & FTN & - & - & 87.14 \\
& GTE (FT) & FTN & - & - & 91.02 \\
& TableFormer & FTN & 97.5 & 96.0 & \textbf{96.8} \\ \bottomrule \rule{0pt}{2.5ex}
& EDD  & TB & 86.0 & - & 86.0 \\ 
& TableFormer & TB & 89.6 & - & \textbf{89.6} \\ \bottomrule \rule{0pt}{3ex}
& TableFormer & STN & 96.9 & 95.7 & 96.7 \\ \bottomrule

\end{tabular}
\caption{\label{tab:structure_results} Structure results on PubTabNet (PTN), FinTabNet (FTN), TableBank (TB) and SynthTabNet (STN). \\ FT: Model was trained on PubTabNet then finetuned.}
\end{table}

\textbf{Cell Detection.} Like any object detector, our \textit{Cell BBox Detector} provides bounding boxes that can be improved with post-processing during inference.
We make use of the grid-like structure of tables to refine the predictions. A detailed explanation on the post-processing is available in the supplementary material. As shown in Tab. \ref{tab:detection}, we evaluate our \textit{Cell BBox Decoder} accuracy for cells with a class label of `content' only using the PASCAL VOC mAP metric for pre-processing and post-processing. Note that we do not have post-processing results for SynthTabNet as images are only provided. To compare the performance of our proposed approach, we've integrated TableFormer's \textit{Cell BBox Decoder} into EDD architecture. As mentioned previously, the Structure Decoder provides the \textit{Cell BBox Decoder} with the features needed to predict the bounding box predictions. Therefore, the accuracy of the \textit{Structure Decoder} directly influences the accuracy of the \textit{Cell BBox Decoder}. If the \textit{Structure Decoder} predicts an extra column, this will result in an extra column of predicted bounding boxes. 

\begin{table}[H]
\begin{tabular}{ccccc}
\toprule
 & Model & Dataset & mAP & mAP (PP) \\\midrule
 & EDD+BBox & PubTabNet & 79.2 & 82.7 \\
 & TableFormer & PubTabNet & \textbf{82.1} & \textbf{86.8} \\
 & TableFormer & SynthTabNet & 87.7 &- \\
\end{tabular}
\caption{\label{tab:detection} Cell Bounding Box detection results on PubTabNet, and FinTabNet. PP: Post-processing. }
\end{table}

\textbf{Cell Content.} In this section, we evaluate the entire pipeline of recovering a table with content. Here we put our approach to test by capitalizing on extracting content from the PDF cells rather than decoding from images.  Tab. \ref{tab:content_result} shows the TEDs score of HTML code representing the structure of the table along with the content inserted in the data cell and compared with the ground-truth. Our method achieved a \textbf{5.3\%} increase over the state-of-the-art, and commercial solutions. We believe our scores would be higher if the HTML ground-truth matched the extracted PDF cell content. Unfortunately, there are small discrepancies such as spacings around words or special characters with various unicode representations.

\begin{table}[H]
\centering
\begin{tabular}{@{}cccccc@{}}
\toprule
& \multirow{2}{*}{Model} & \multicolumn{3}{c}{TEDS} \\
& & Simple & Complex  & All \\ \midrule
& Tabula & 78.0 & 57.8 & 67.9 \\
& Traprange & 60.8 & 49.9 & 55.4 \\
& Camelot  & 80.0 & 66.0 & 73.0 \\
& Acrobat$^{\textregistered}$ Pro  & 68.9 & 61.8 & 65.3 \\
& EDD & 91.2 & 85.4 & 88.3 \\
& TableFormer & 95.4 & 90.1 & \textbf{93.6} \\

\end{tabular}
\caption{\label{tab:content_result} Results of structure with content retrieved using cell detection on PubTabNet. In all cases the input is PDF documents with cropped tables.}
\end{table}

\begin{figure*}[hbt!]
	\centering
	\includegraphics[scale=0.27]{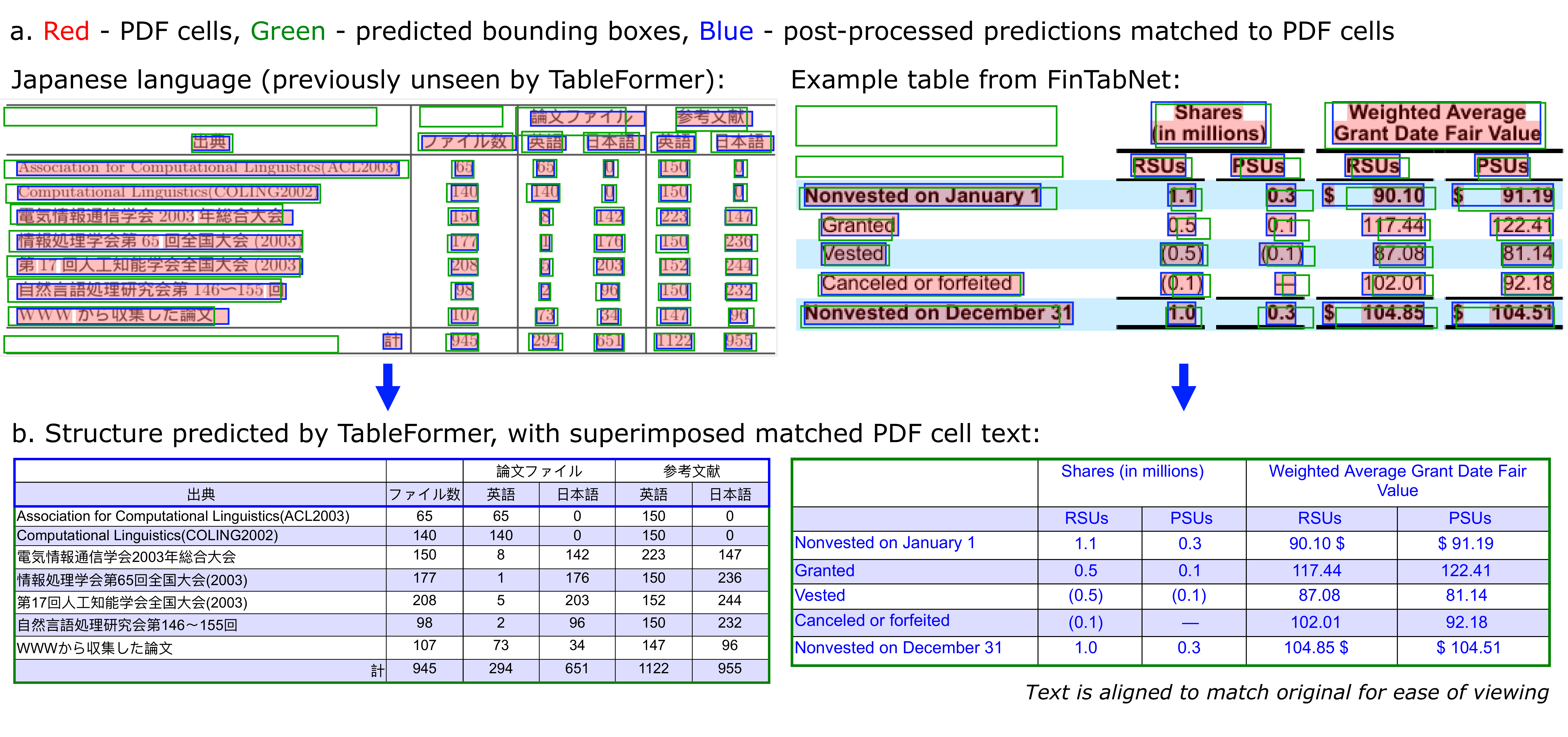}
	\caption{\label{fig:predictions1} One of the benefits of TableFormer is that it is language agnostic, as an example, the left part of the illustration demonstrates TableFormer predictions on previously unseen language (Japanese). Additionally, we see that TableFormer is robust to variability in style and content, right side of the illustration shows the example of the TableFormer prediction from the FinTabNet dataset.}
\end{figure*}

\begin{figure*}[hbt!]
	\centering
	\includegraphics[scale=0.60]{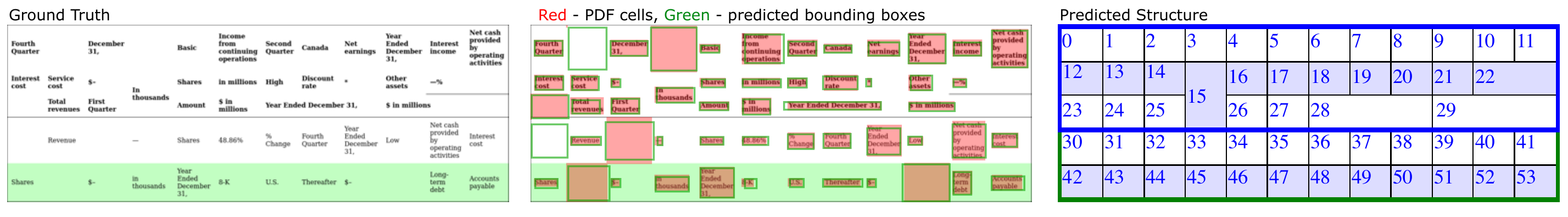}
	\caption{\label{fig:predictions2} An example of TableFormer predictions (bounding boxes and structure) from generated SynthTabNet table.}
\end{figure*}

\subsection{Qualitative Analysis}
We showcase several visualizations for the different components of our network on various \textit{``complex"} tables within datasets presented in this work in Fig. \ref{fig:predictions1} and Fig. \ref{fig:predictions2}
As it is shown, our model is able to predict bounding boxes for all table cells, even for the empty ones. Additionally, our post-processing techniques can extract the cell content by matching the predicted bounding boxes to the PDF cells based on their overlap and spatial proximity.
The left part of Fig. \ref{fig:predictions1} demonstrates also the adaptability of our method to any language, as it can successfully extract Japanese text, although the training set contains only English content.
We provide more visualizations including the intermediate steps in the supplementary material.
Overall these illustrations justify the versatility of our method across a diverse range of table appearances and content type.

\section{Future Work \& Conclusion}
In this paper, we presented TableFormer an end-to-end transformer based approach to predict table structures and bounding boxes of cells from an image.
This approach enables us to recreate the table structure, and extract the cell content from PDF or OCR by using bounding boxes.
Additionally, it provides the versatility required in real-world scenarios when dealing with various types of PDF documents, and languages.
Furthermore, our method outperforms all state-of-the-arts with a wide margin.
Finally, we introduce ``SynthTabNet" a challenging synthetically generated dataset that reinforces missing characteristics from other datasets.

{\small
\bibliographystyle{ieee_fullname}
\bibliography{egbib}
}

\setcounter{section}{0}

\clearpage
\begin{strip}
\begin{center}

\textbf{\large{ TableFormer: Table Structure Understanding with Transformers \\ \myfont{Supplementary Material}}}

\end{center}
\end{strip}
\section{\label{supsec:Intro}Details on the datasets}

\subsection{Data preparation}

\begin{figure*}[htb!]\centering
\centering
\includegraphics[scale=0.17]{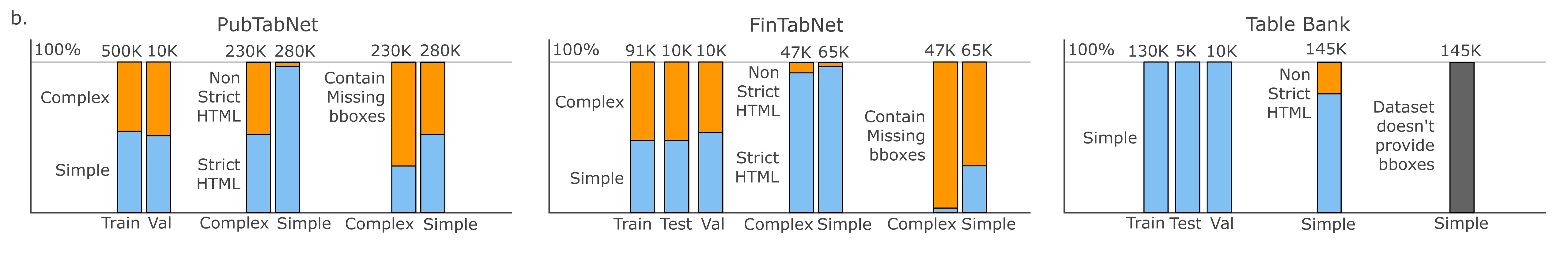}
    \caption{\label{supfig:Fig1} Distribution of the tables across different dimensions per dataset.
    Simple vs complex tables per dataset and split, strict vs non strict html structures per dataset
    and table complexity, missing bboxes per dataset and table complexity.}
\end{figure*}

As a first step of our data preparation process, we have calculated statistics over the datasets across the following dimensions:
(1) table size measured in the number of rows and columns,
(2) complexity of the table,
(3) strictness of the provided HTML structure and
(4) completeness (i.e. no omitted bounding boxes).
A table is considered to be simple if it does not contain row spans or column spans.
Additionally, a table has a strict HTML structure if every row has the same number of columns after taking into account any row or column spans.
Therefore a strict HTML structure looks always rectangular.
However, HTML is a lenient encoding format, i.e. tables with rows of different sizes might still be regarded as correct due to implicit display rules.
These implicit rules leave room for ambiguity, which we want to avoid.
As such, we prefer to have "strict" tables, i.e. tables where every row has exactly the same length.

We have developed a technique that tries to derive a missing bounding box out of its neighbors.
As a first step, we use the annotation data to generate the most fine-grained grid that covers the table structure.
In case of strict HTML tables, all grid squares are associated with some table cell and in the presence of table spans a cell extends across multiple grid squares.
When enough bounding boxes are known for a rectangular table, it is possible to compute the geometrical border lines between the grid rows and columns.
Eventually this information is used to generate the missing bounding boxes.
Additionally, the existence of unused grid squares indicates that the table rows have unequal number of columns and the overall structure is non-strict.
The generation of missing bounding boxes for non-strict HTML tables is ambiguous and therefore quite challenging.
Thus, we have decided to simply discard those tables.
In case of PubTabNet we have computed missing bounding boxes for 48\% of the simple and 69\% of the complex tables.
Regarding FinTabNet, 68\% of the simple and 98\% of the complex tables require the generation of bounding boxes.

Figure~\ref{supfig:Fig1} illustrates the distribution of the tables across different dimensions per
dataset.

\subsection{Synthetic datasets}

Aiming to train and evaluate our models in a broader spectrum of table data we have synthesized four types of datasets.
Each one contains tables with different appearances in regard to their size, structure, style and content.
Every synthetic dataset contains 150k examples, summing up to 600k synthetic examples.
All datasets are divided into Train, Test and Val splits (80\%, 10\%, 10\%).

The process of generating a synthetic dataset can be decomposed into the following steps:

1. Prepare styling and content templates:
The styling templates have been manually designed and organized into groups of scope specific appearances (e.g. financial data, marketing data, etc.)
Additionally, we have prepared curated collections of content templates by extracting the most frequently used terms out of non-synthetic datasets (e.g. PubTabNet, FinTabNet, etc.).

2. Generate table structures:
The structure of each synthetic dataset assumes a horizontal table header which potentially spans over multiple rows and a table body that may contain a combination of row spans and column spans.
However, spans are not allowed to cross the header - body boundary.
The table structure is described by the parameters:
Total number of table rows and columns, number of header rows, type of spans (header only spans, row only spans, column only spans, both row and column spans), maximum span size and the ratio of the table area covered by spans.

3. Generate content:
Based on the dataset \textit{theme}, a set of suitable content templates is chosen first.
Then, this content can be combined with purely random text to produce the synthetic content.

4. Apply styling templates:
Depending on the domain of the synthetic dataset, a set of styling templates is first manually selected.
Then, a style is randomly selected to format the appearance of the synthesized table.

5. Render the complete tables:
The synthetic table is finally rendered by a web browser engine to generate the bounding boxes for each table cell.
A batching technique is utilized to optimize the runtime overhead of the rendering process.

\section{\label{supsec:Intro}Prediction post-processing for PDF documents}


Although TableFormer can predict the table structure and the bounding boxes for tables recognized inside PDF documents,
this is not enough when a full reconstruction of the original table is required. 
This happens mainly due the following reasons:

\begin{itemize}
    \item TableFormer output does not include the table cell content.
    \item There are occasional inaccuracies in the predictions of the bounding boxes.
\end{itemize}

However, it is possible to mitigate those limitations by combining the TableFormer predictions with the information already present inside a programmatic PDF document.
More specifically, PDF documents can be seen as a sequence of PDF cells where each cell is described by its content and bounding box.
If we are able to associate the PDF cells with the predicted table cells, we can directly link the PDF cell content to the table cell structure and
use the PDF bounding boxes to correct misalignments in the predicted table cell bounding boxes.

Here is a step-by-step description of the prediction post-processing:

1. Get the minimal grid dimensions - number of rows and columns for the predicted table structure.
This represents the most granular grid for the underlying table structure.

2. Generate pair-wise matches between the bounding boxes of the PDF cells and the predicted cells.
The Intersection Over Union (IOU) metric is used to evaluate the quality of the matches.

3. Use a carefully selected IOU threshold to designate the matches as ``good'' ones and ``bad'' ones.

3.a. If all IOU scores in a column are below the threshold, discard all predictions (structure and bounding boxes) for that column.

4. Find the best-fitting content alignment for the predicted cells with good IOU per each column.
The alignment of the column can be identified by the following formula:

\label{eq:loss}
\begin{equation}
\begin{split}
    alignment = \argmin_c\{D_c\} \\
    D_c = max\{x_c\} - min\{x_c\}
\end{split}
\end{equation}
where $c$ is one of \{left, centroid, right\} and $x_c$ is the x-coordinate for the corresponding point.



5. Use the alignment computed in step 4, to compute the median $x$-coordinate for all table columns and the median cell size for all table cells.
The usage of median during the computations, helps to eliminate outliers caused by occasional column spans which are usually wider than the normal.

6. Snap all cells with bad IOU to their corresponding median $x$-coordinates and cell sizes.

7. Generate a new set of pair-wise matches between the corrected bounding boxes and PDF cells.
This time use a modified version of the IOU metric, where the area of the intersection between the predicted and PDF cells is divided by the PDF cell area.
In case there are multiple matches for the same PDF cell, the prediction with the higher score is preferred.
This covers the cases where the PDF cells are smaller than the area of predicted or corrected prediction cells.

8. In some rare occasions, we have noticed that TableFormer can confuse a single column as two.
When the post-processing steps are applied, this results with two predicted columns pointing to the same PDF column.
In such case we must de-duplicate the columns according to highest total column intersection score.

9. Pick up the remaining orphan cells. 
There could be cases, when after applying all the previous post-processing steps, some PDF cells could still remain without any match to predicted cells.
%
%
%
%
However, it is still possible to deduce the correct matching for an orphan PDF cell by mapping its bounding box on the geometry of the grid.
This mapping decides if the content of the orphan cell will be appended to an already matched table cell, or a new table cell should be created to match with the orphan.

9a. Compute the top and bottom boundary of the horizontal band for each grid row (min/max $y$ coordinates per row).

9b. Intersect the orphan's bounding box with the row bands, and map the cell to the closest grid row.

9c. Compute the left and right boundary of the vertical band for each grid column (min/max $x$ coordinates per column).

9d. Intersect the orphan's bounding box with the column bands, and map the cell to the closest grid column.

9e. If the table cell under the identified row and column is not empty, extend its content with the content of the orphan cell.

9f. Otherwise create a new structural cell and match it wit the orphan cell.

\textit{Aditional images with examples of TableFormer predictions and post-processing can be found below.}

\begin{figure}[htb!]\centering
\centering
\includegraphics[scale=0.37]{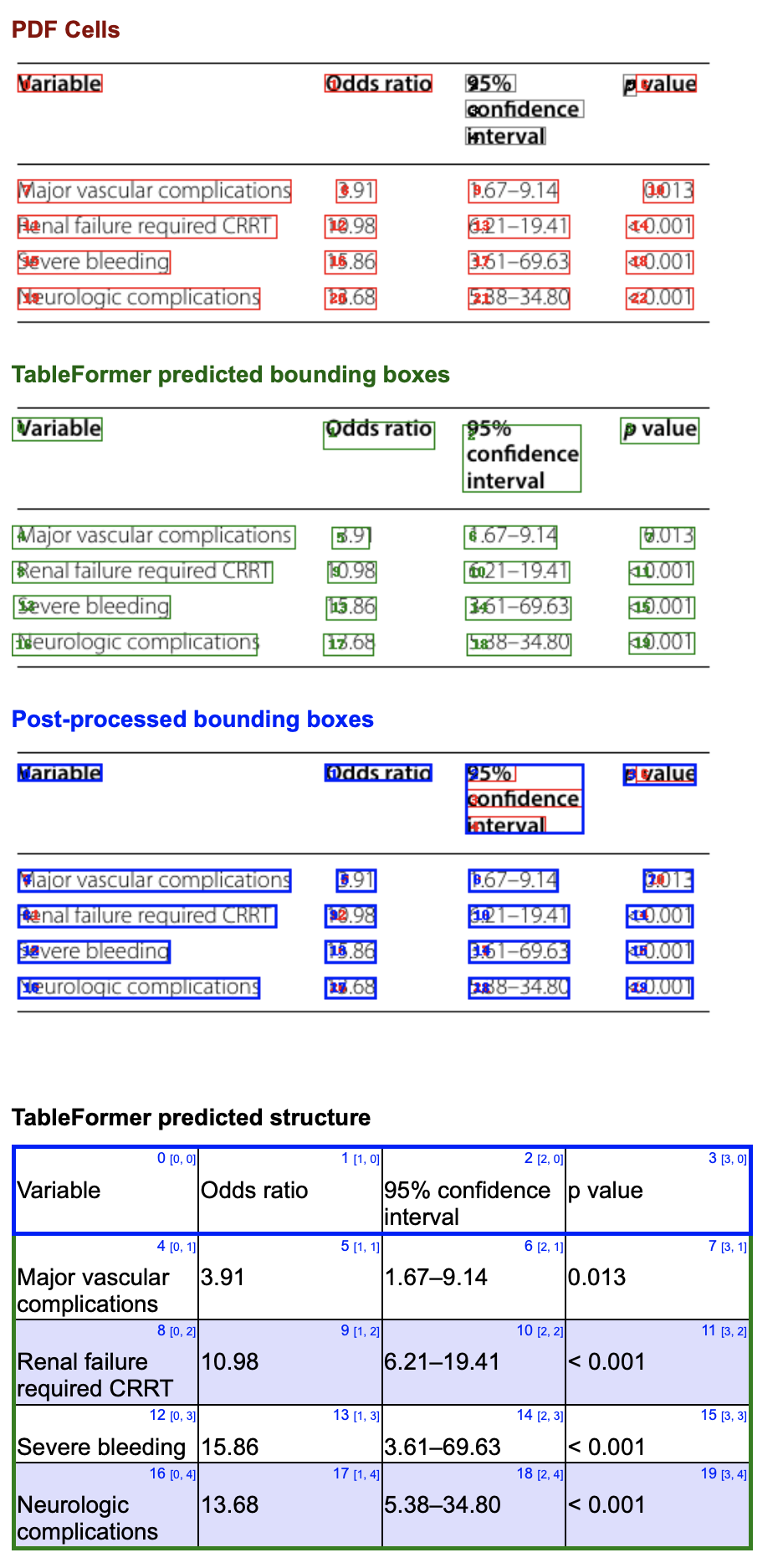}
    \caption{\label{supfig:Fig2} Example of a table with multi-line header.}
\end{figure}

\begin{figure}[htb!]\centering
\centering
\includegraphics[scale=0.23]{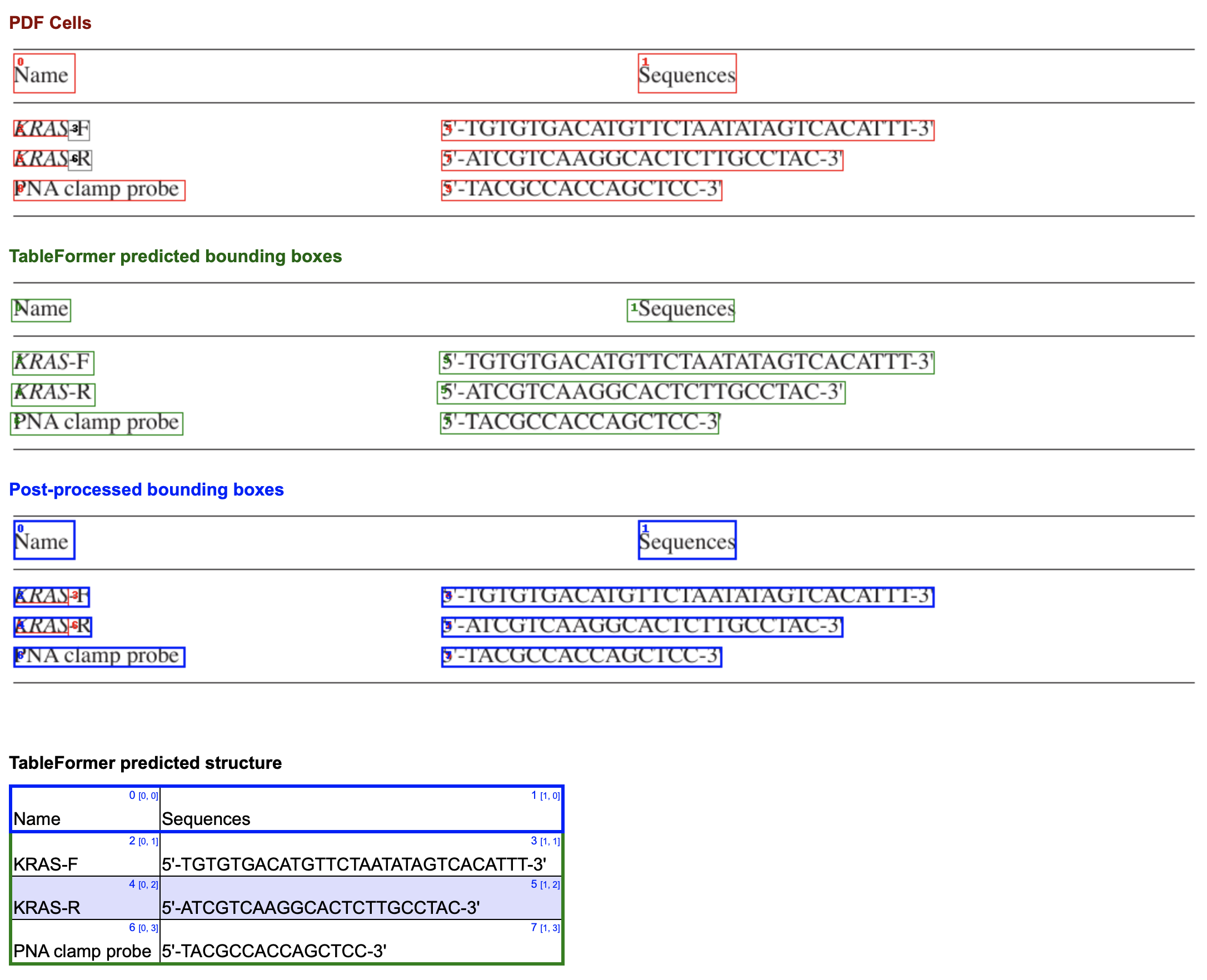}
    \caption{\label{supfig:Fig3} Example of a table with big empty distance between cells.}
\end{figure}

\begin{figure}[htb!]\centering
\centering
\includegraphics[scale=0.33]{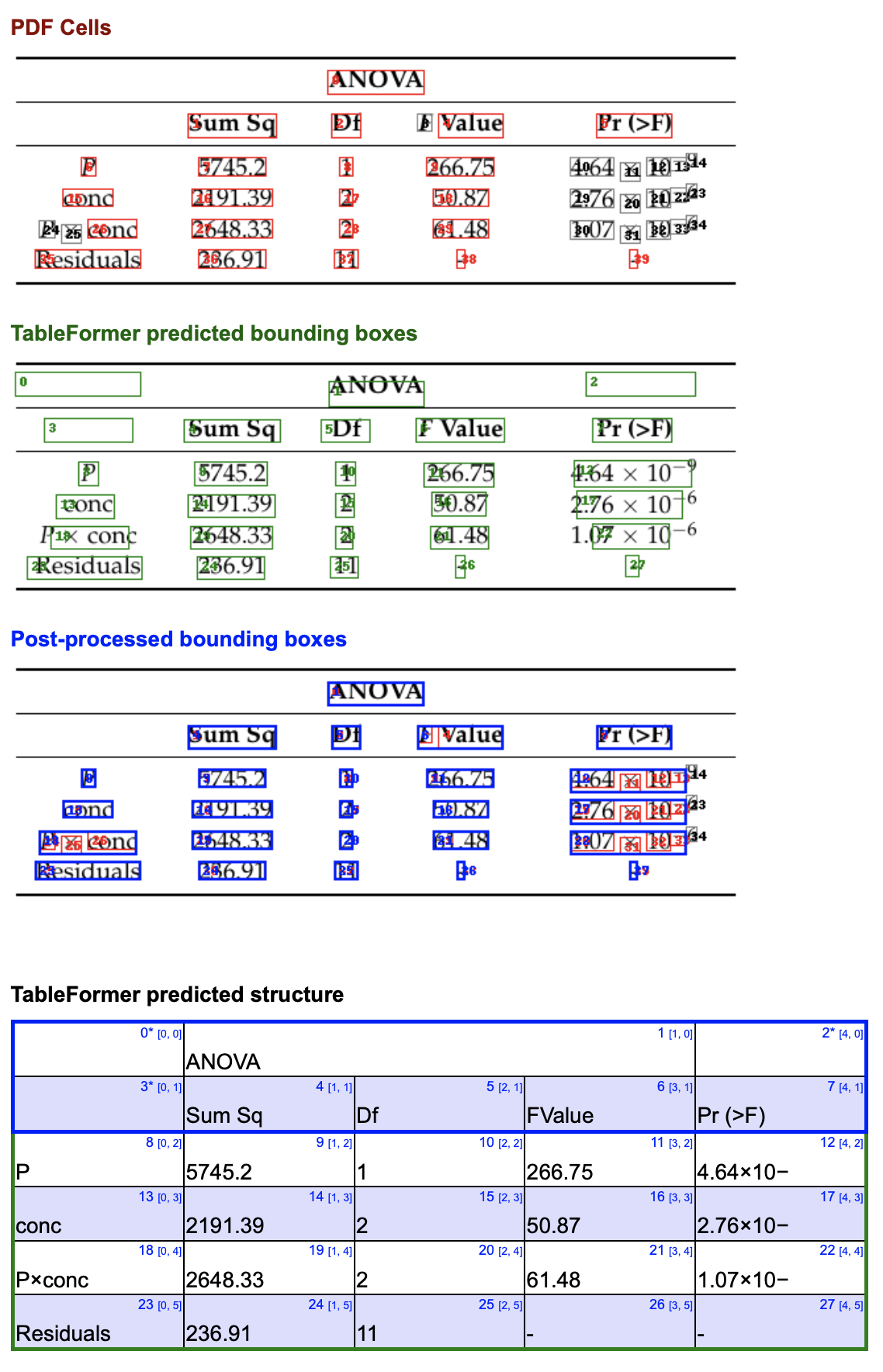}
    \caption{\label{supfig:Fig4} Example of a complex table with empty cells.}
\end{figure}

\begin{figure}[htb!]\centering
\centering
\includegraphics[scale=0.27]{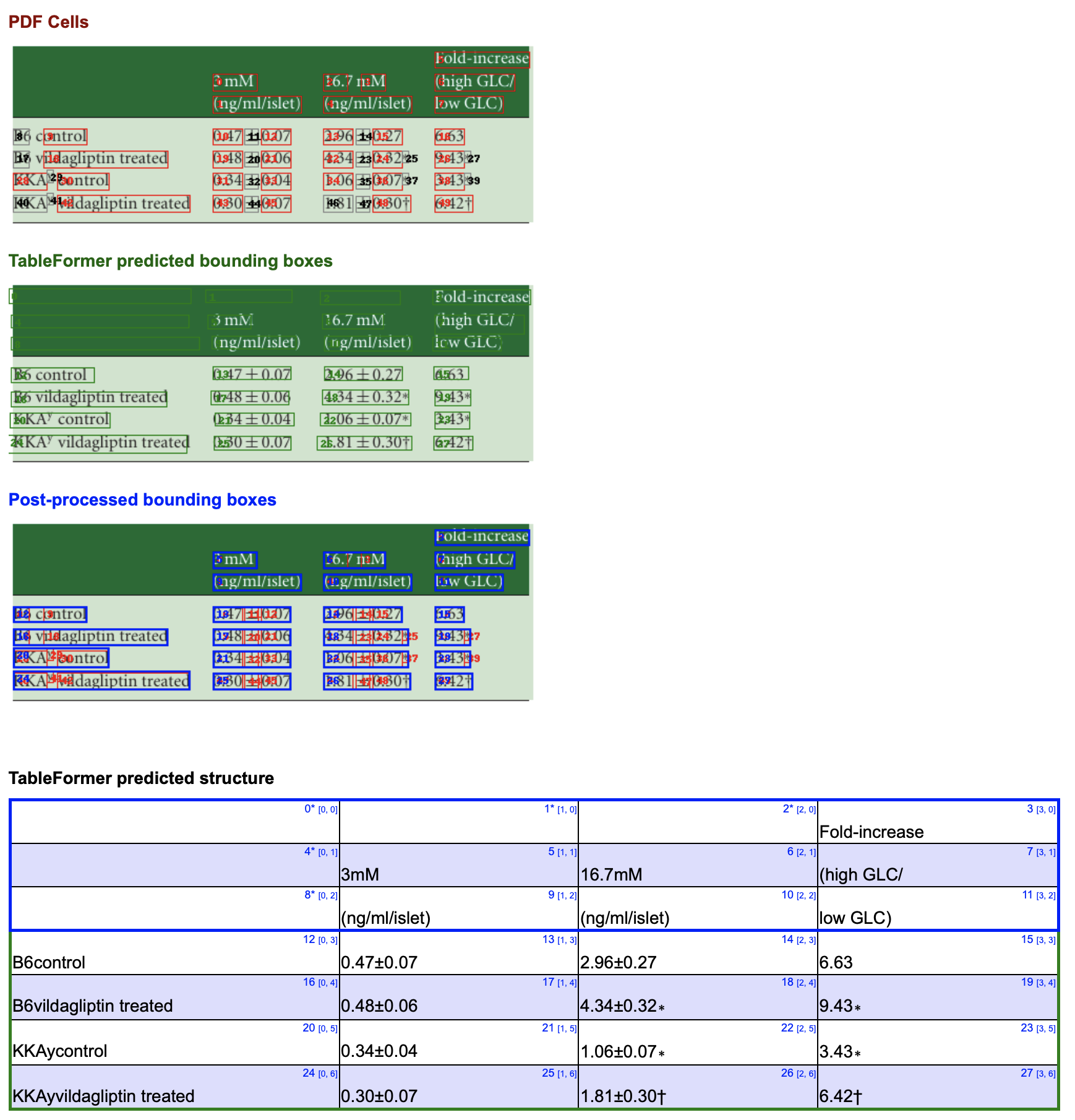}
    \caption{\label{supfig:Fig5} Simple table with different style and empty cells.}
\end{figure}

\begin{figure}[htb!]\centering
\centering
\includegraphics[scale=0.24]{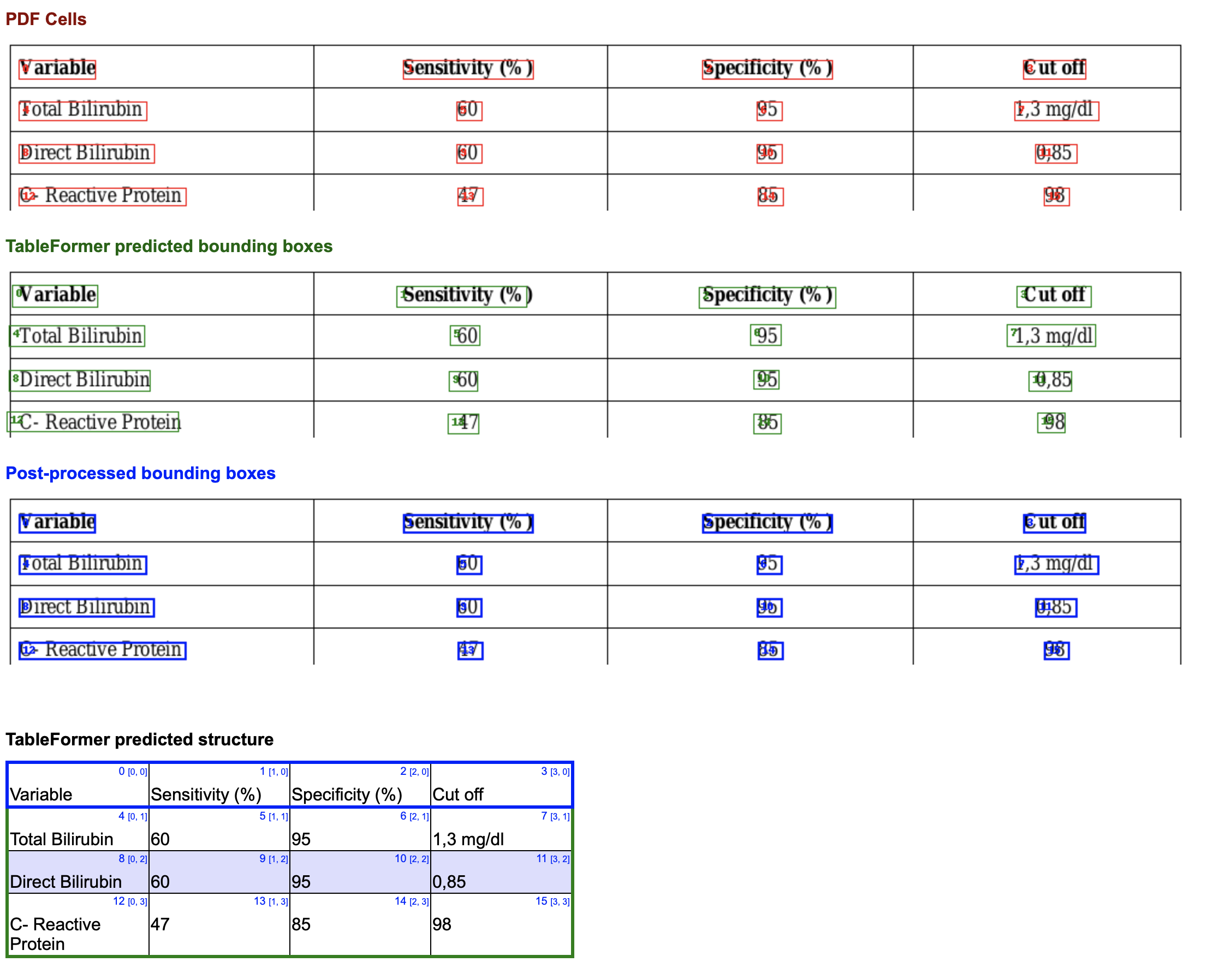}
    \caption{\label{supfig:Fig6} Simple table predictions and post processing.}
\end{figure}

\begin{figure}[htb!]\centering
\centering
\includegraphics[scale=0.25]{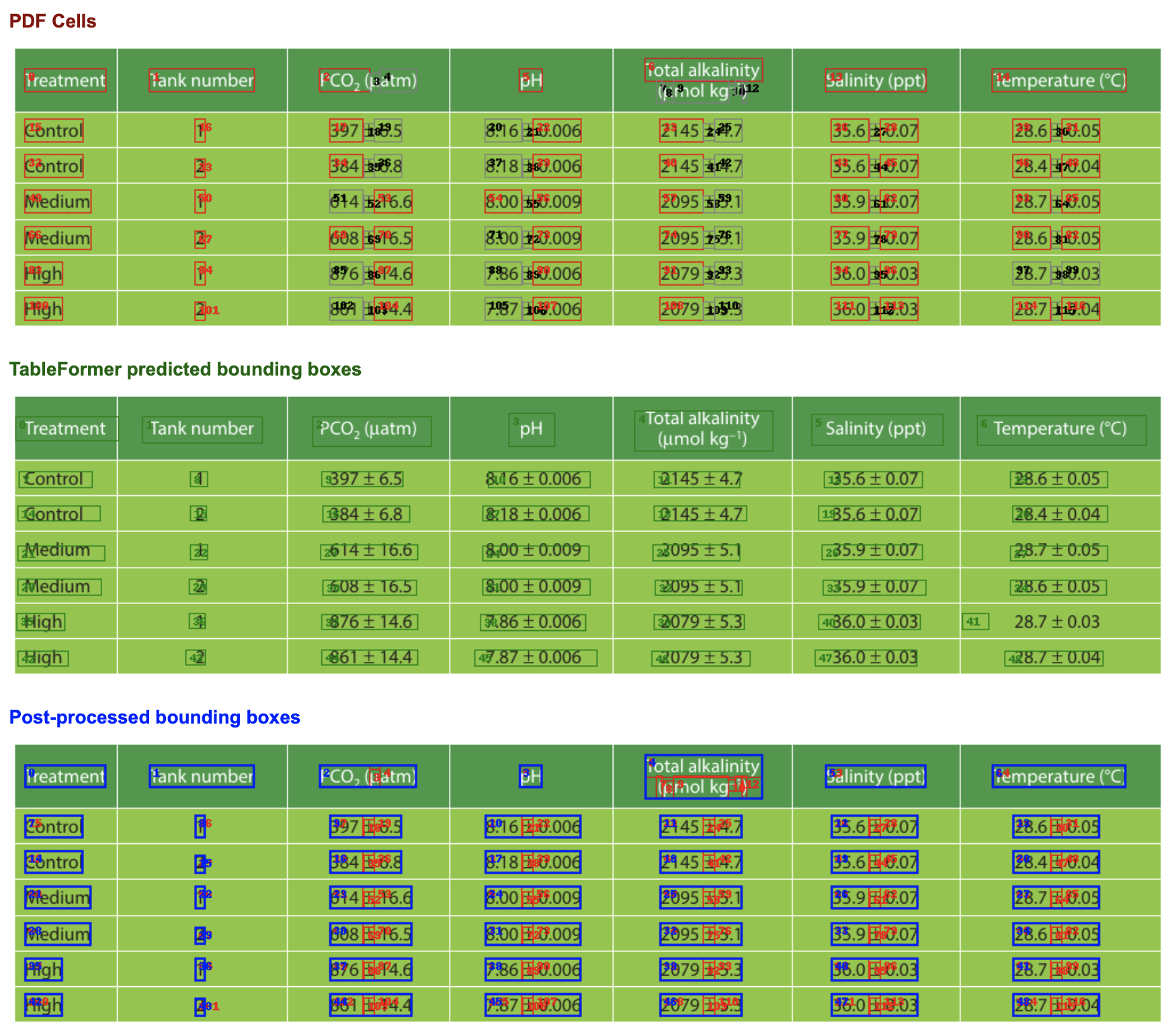}
\includegraphics[scale=0.25]{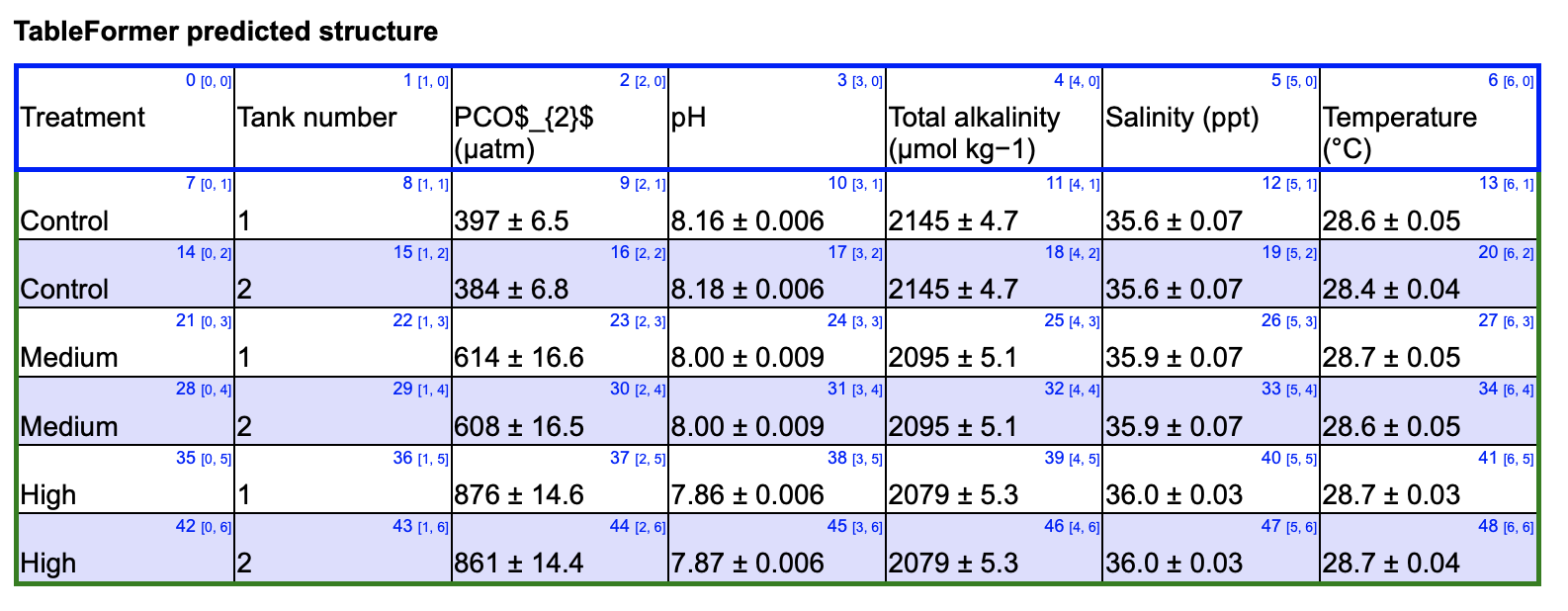}
    \caption{\label{supfig:Fig7} Table predictions example on colorful table.}
\end{figure}

\begin{figure}[ht!]\centering
\centering
\includegraphics[scale=0.25]{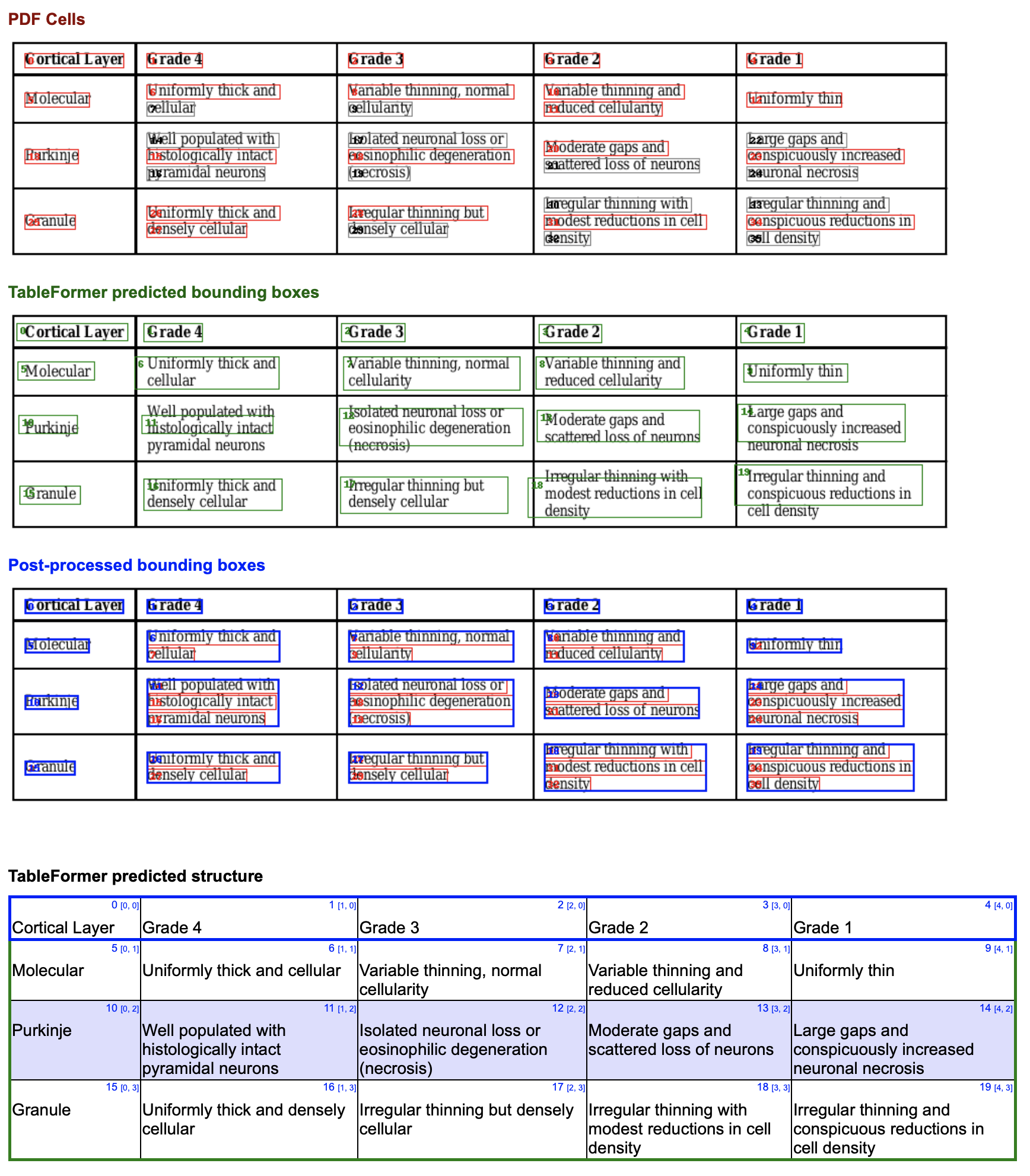}
    \caption{\label{supfig:Fig8} Example with multi-line text.}
\end{figure}

\begin{figure}[htb!]\centering
\centering
\includegraphics[scale=0.26]{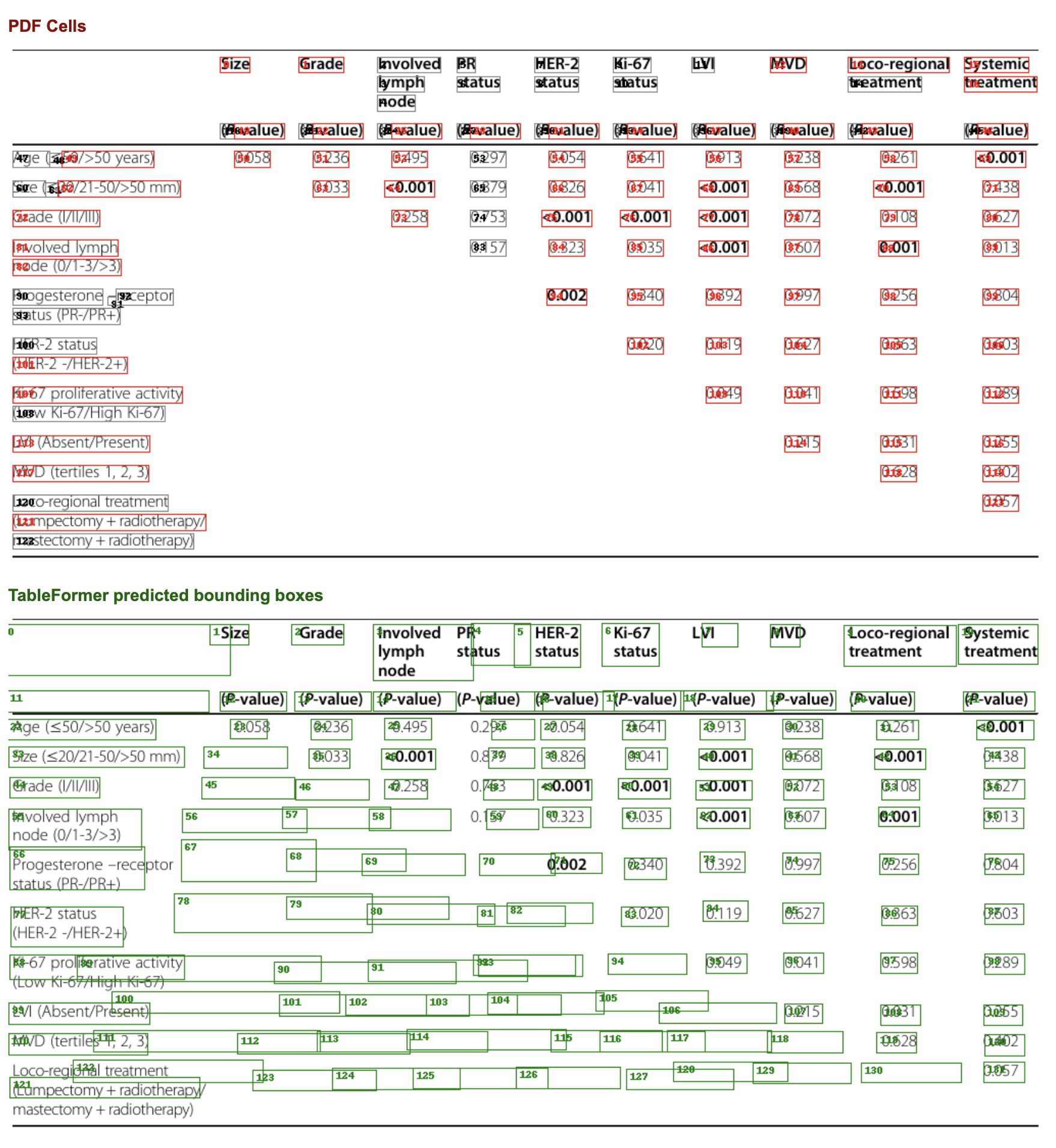}
\includegraphics[scale=0.26]{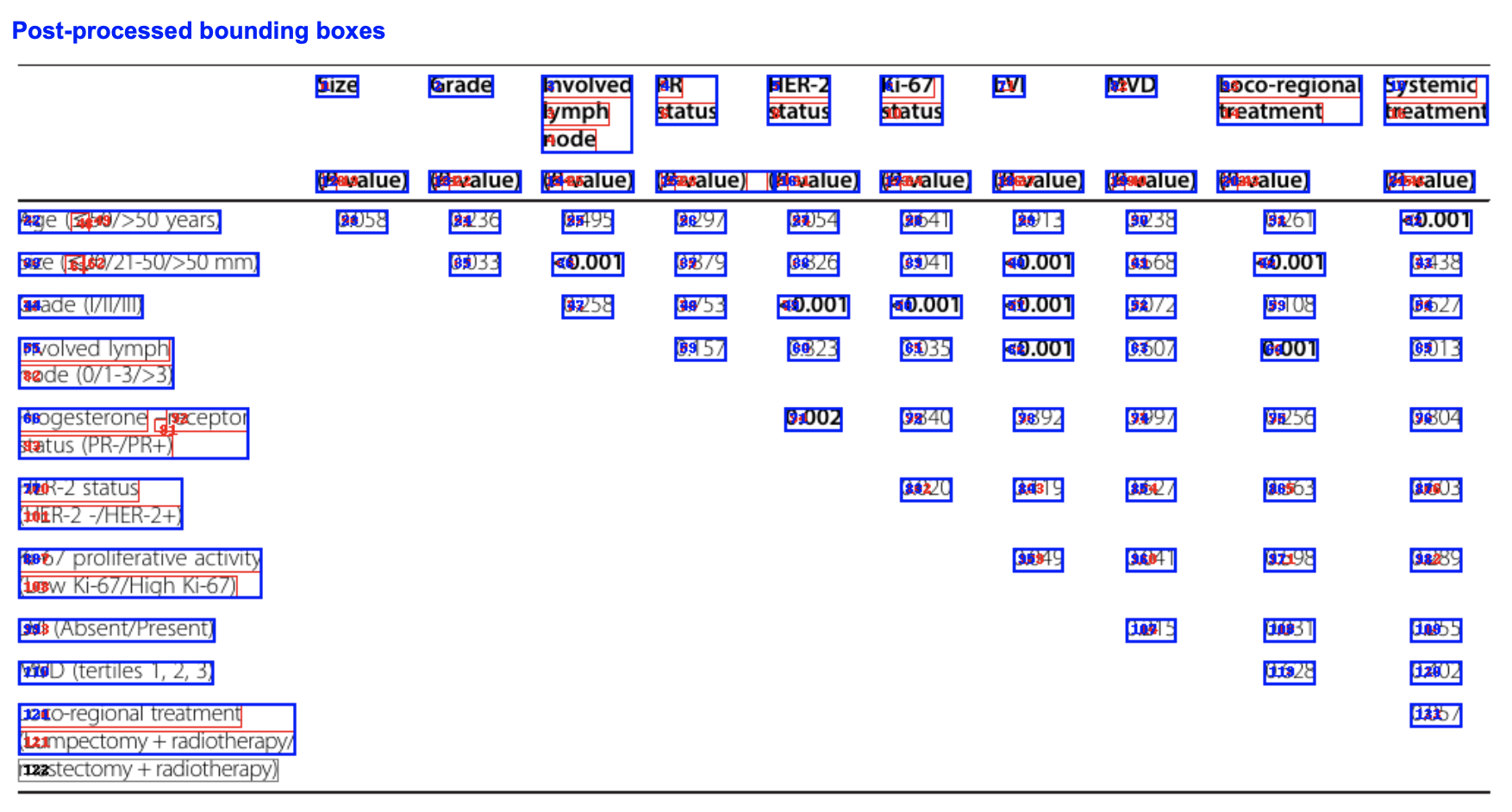}
\includegraphics[scale=0.21]{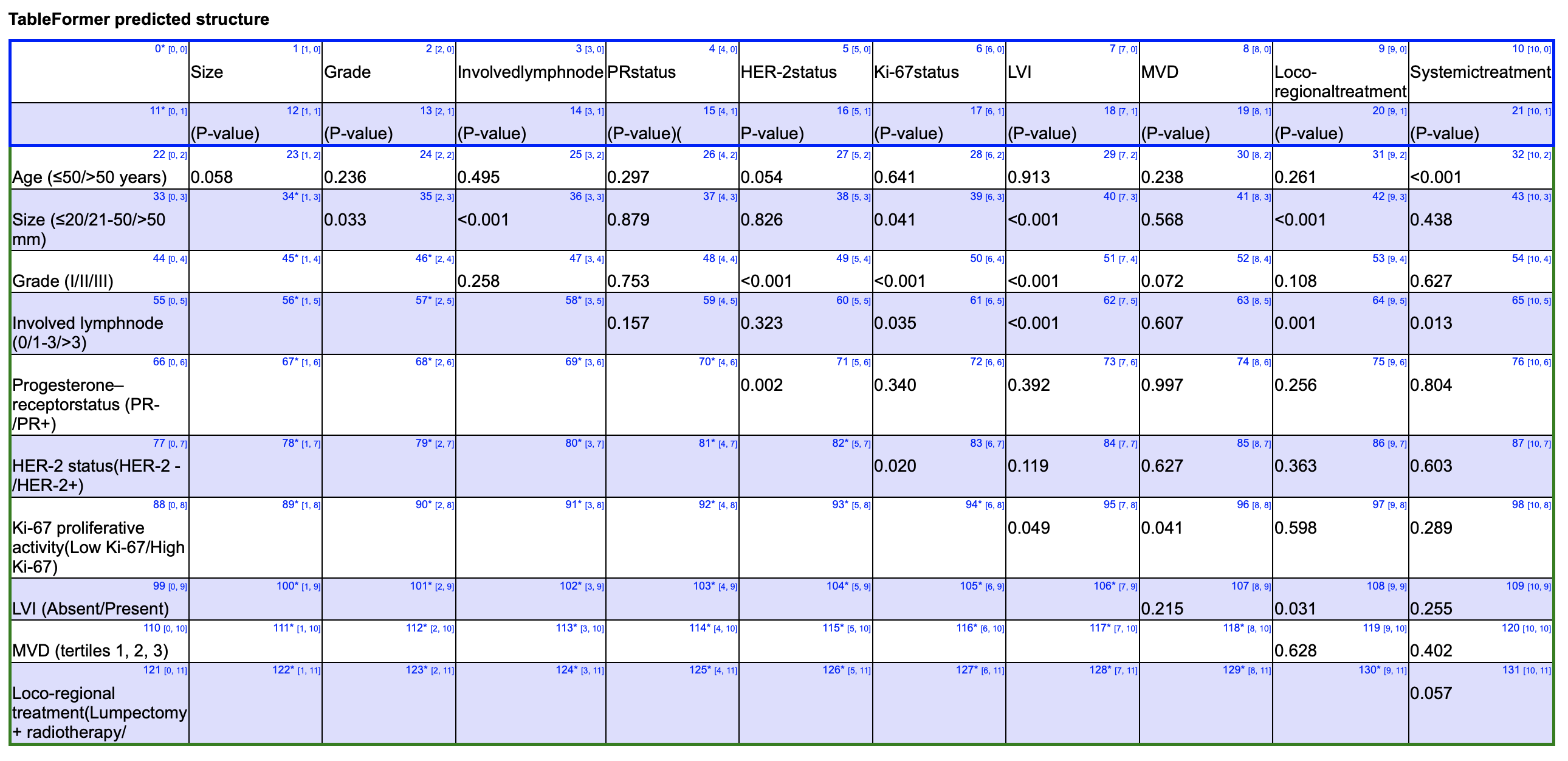}
    \caption{\label{supfig:Fig9} Example with triangular table.}
\end{figure}

\begin{figure}[htb!]\centering
\centering
\includegraphics[scale=0.41]{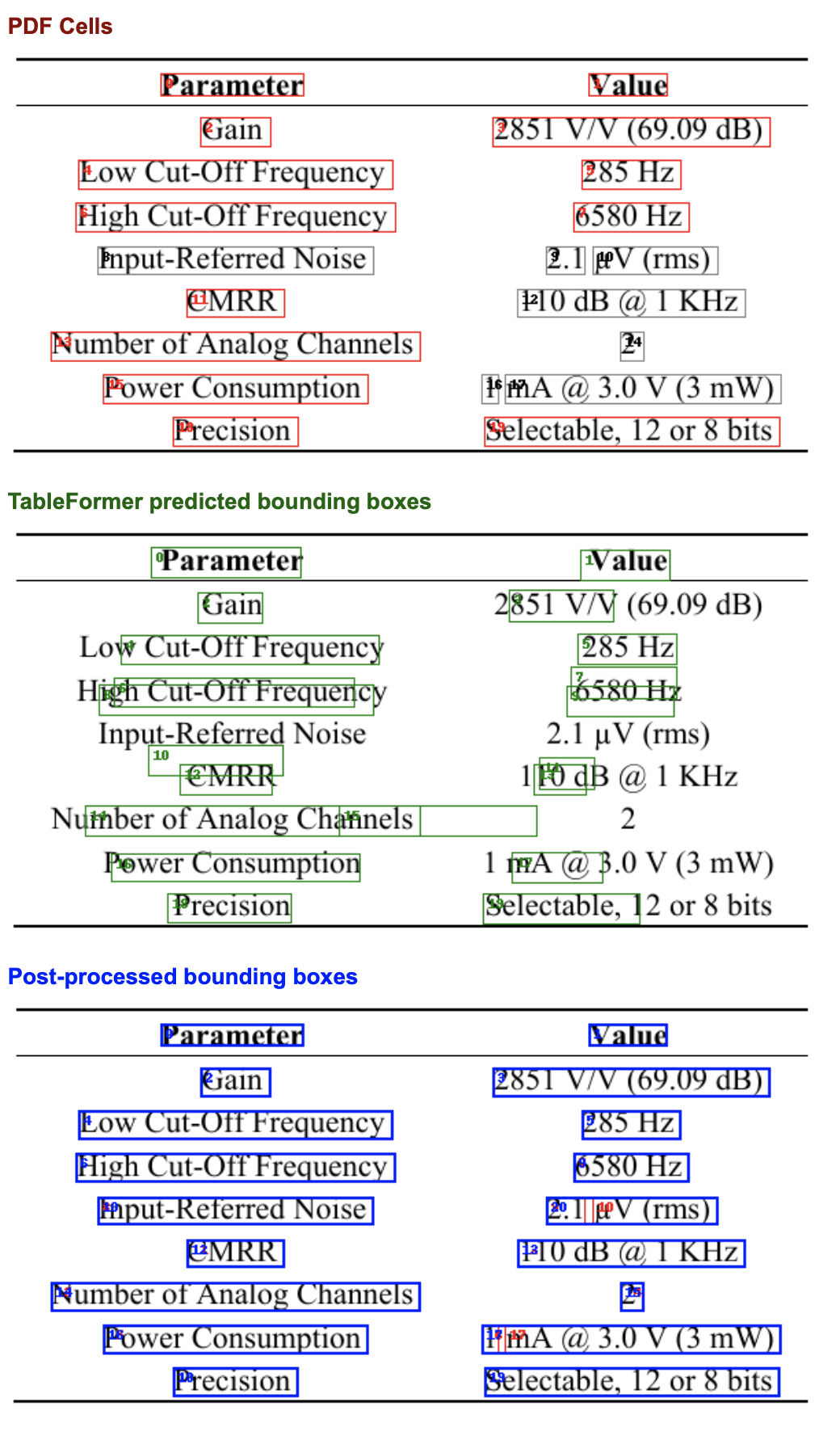}
\includegraphics[scale=0.41]{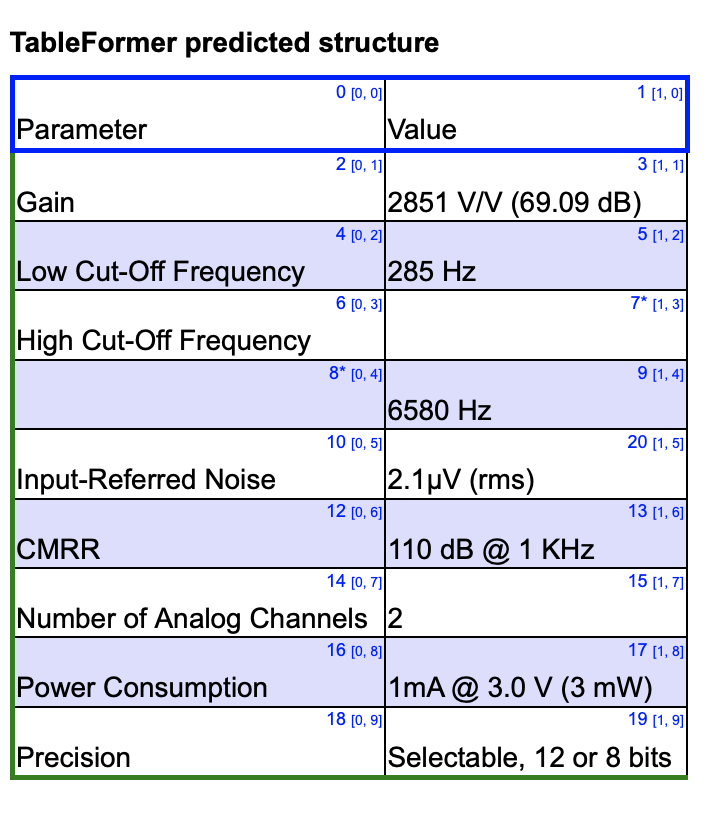}
    \caption{\label{supfig:Fig10} Example of how post-processing helps to restore mis-aligned bounding boxes prediction artifact.}
\end{figure}

\begin{figure*}[ht]\centering
\centering
\includegraphics[scale=0.26]{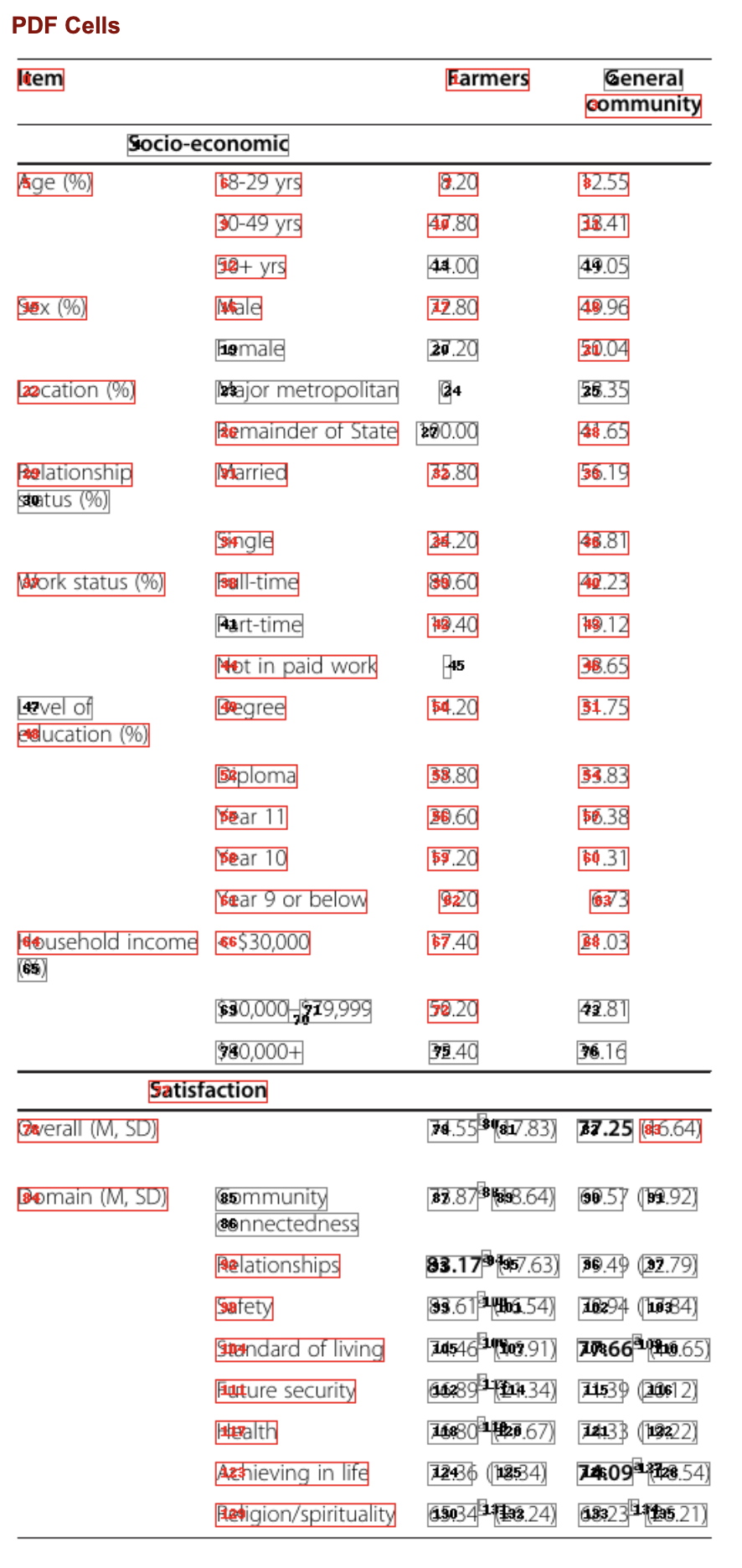}
\includegraphics[scale=0.26]{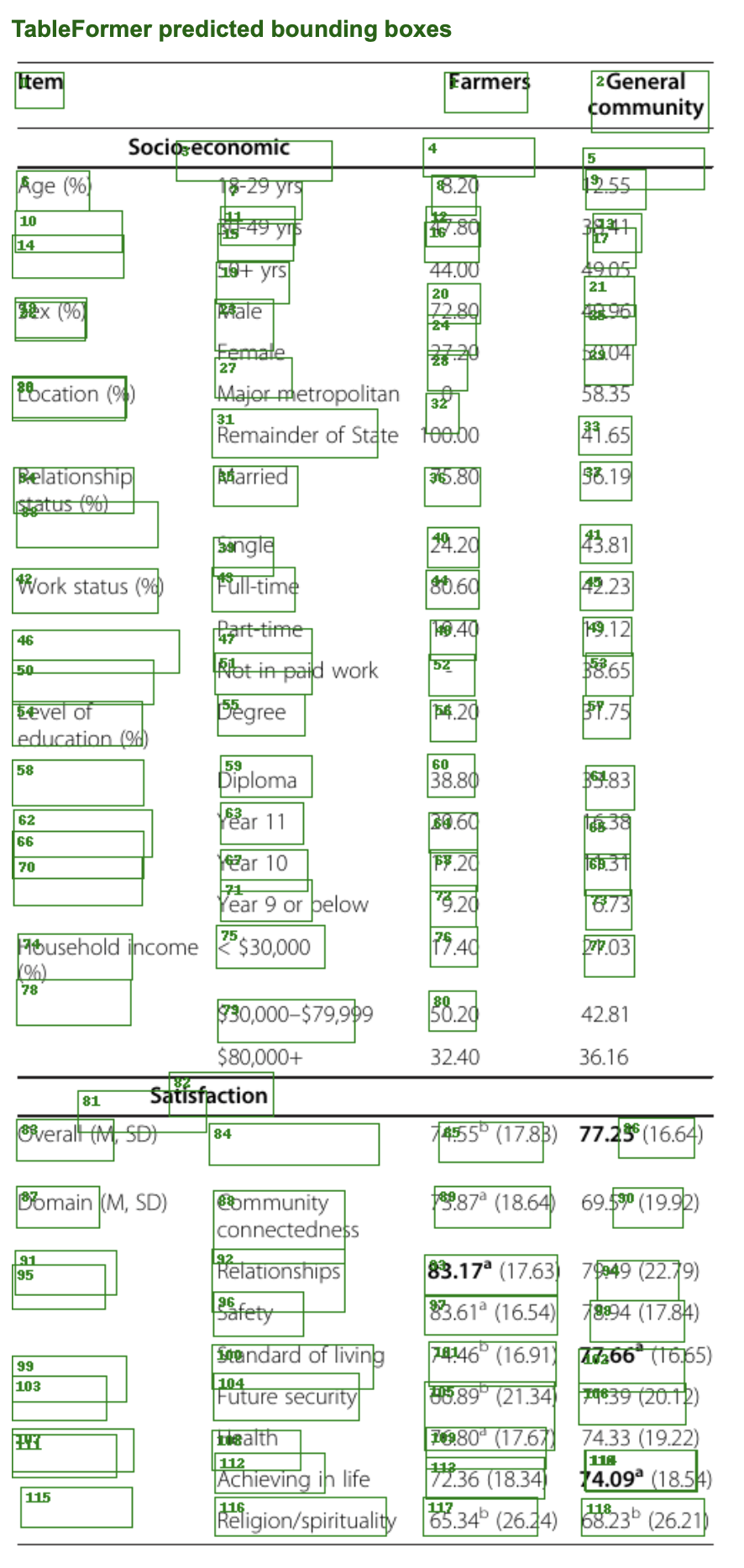}
\includegraphics[scale=0.26]{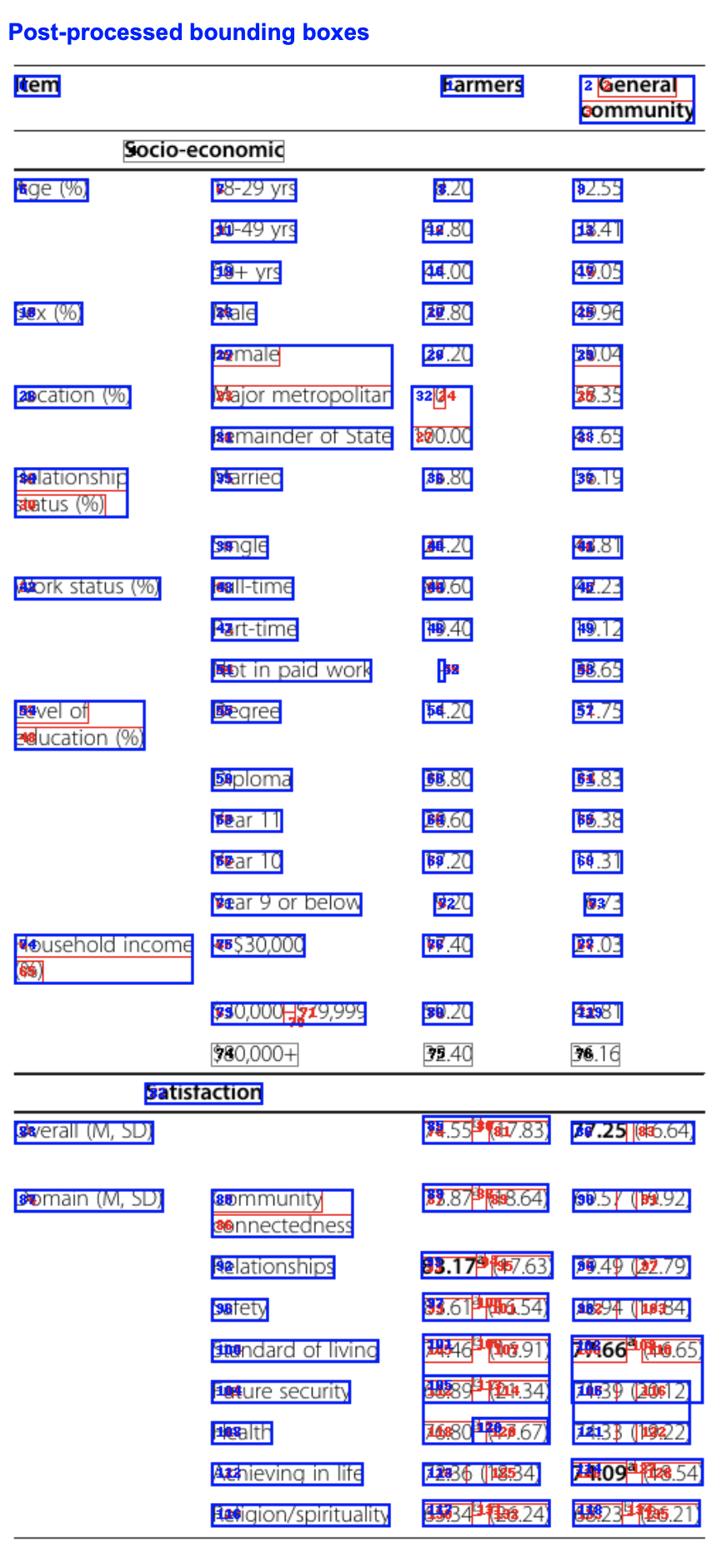}
\includegraphics[scale=0.24]{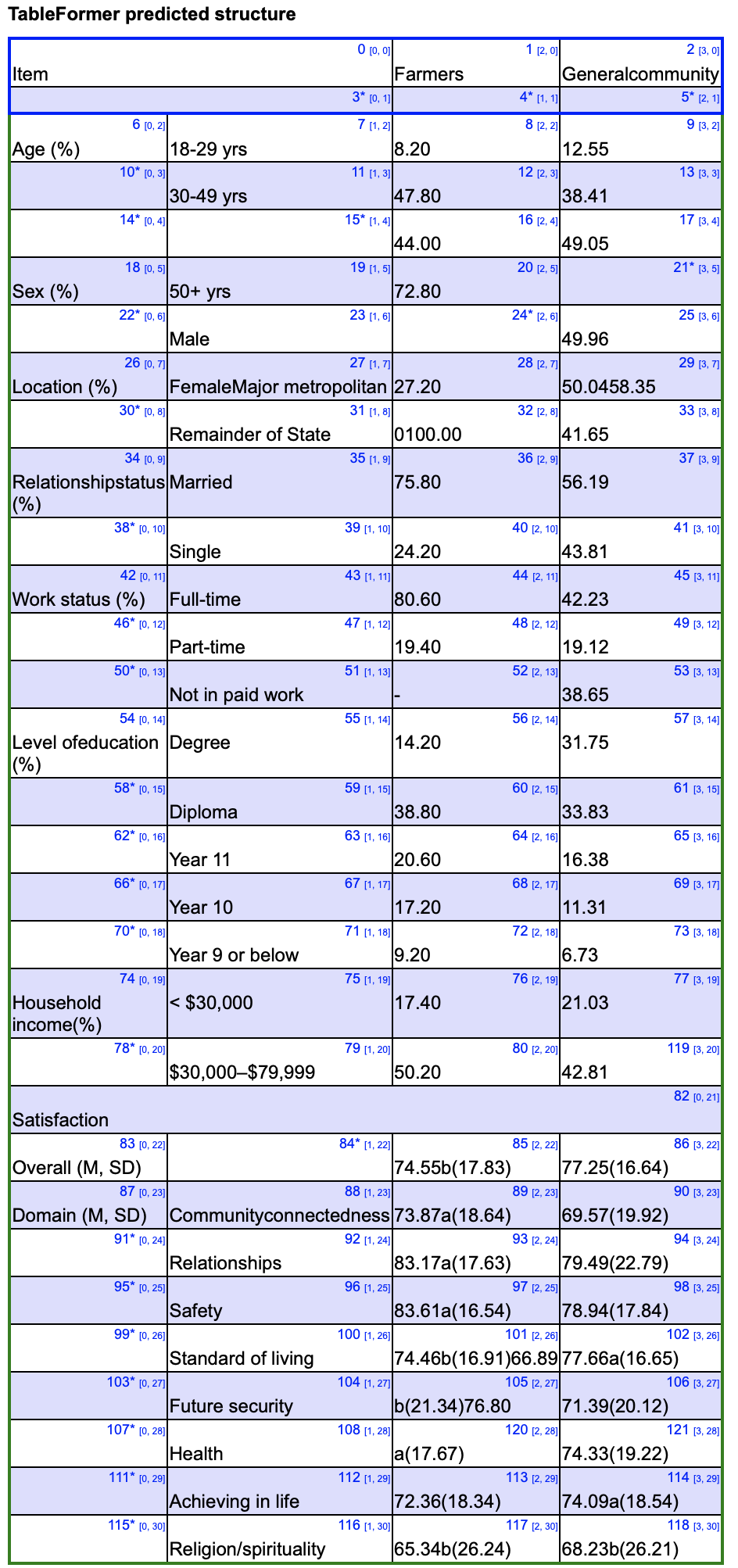}
    \caption{\label{supfig:Fig11} Example of long table. End-to-end example from initial PDF cells to prediction of bounding boxes, post processing and prediction of structure.}
\end{figure*}

\end{document}